\documentclass[twocolumn,twoside]{article}

\usepackage{IJCISIM,multicol,times}

\usepackage{epsfig}

\usepackage{moreverb}
\usepackage{url}
\usepackage{graphicx}
\usepackage{subcaption}
\usepackage{amsmath}
\usepackage{graphicx}
\usepackage{algorithm}
\usepackage{algpseudocode}
\usepackage{multirow}
\usepackage{caption}
\usepackage{subcaption}
\usepackage{diagbox}

\PUBYEAR{2022}
\VOL{14}
\newcommand\startpage{28}
\usepackage{fancyhdr}
\usepackage{ifthen}
\usepackage{lastpage}
\fancypagestyle{mypagestyle}{%
    \fancyhf{} 
    \fancyhead[RO]{Y. Drias et al. }
    \fancyhead[LO]{\thepage}
    \fancyhead[LE]{Enhanced Elephant Herding Optimization for Large Scale Information Access on Social Media}
    \fancyhead[RE]{\thepage}
    \renewcommand{\headrulewidth}{0pt}
}
\begin{document}
\fancypagestyle{plain}{%
  \renewcommand{\headrulewidth}{0pt}%
  \fancyhf{}%
  \fancyfoot[R]{MIR Labs, USA}
}
\sloppy

\title{
\begin{small}Submitted: 28 October, 2021; Accepted: 2 February, 2022; Published: 9 April, 2022\end{small}
\\Enhanced Elephant Herding Optimization for Large Scale Information Access on Social Media}

\author{{\bf Yassine Drias$^1$, Habiba Drias$^2$ and Ilyes Khennak$^3$} \\[1em]
	$^1$University of Algiers, \\
	02 rue Didouche Mourad, Algiers 16000, Algeria \\
	\textit{y.drias@univ-alger.dz} \\[1em]
	$^2$LRIA, USTHB,\\
	BP 32 El Alia, Bab Ezzouar, Algiers 16111, Algeria\\
	\textit{habiba.drias@usthb.edu.dz} \\[1em]
	$^3$LRIA, USTHB,\\
	BP 32 El Alia, Bab Ezzouar, Algiers 16111, Algeria\\
	\textit{Ilyes.khennak@usthb.edu.dz} \\[1em]}

\maketitle

\begin{abstract} In this article, we present a novel information access approach inspired by the information foraging theory (IFT) and elephant herding optimization (EHO). First, we propose a model for information access on social media based on the IFT. We then elaborate an adaptation of the original EHO algorithm to apply it to the information access problem. The combination of the IFT and EHO constitutes a good opportunity to find relevant information on social media. However, when dealing with voluminous data, the performance undergoes a sharp drop. To overcome this issue, we developed an enhanced version of EHO for large scale information access. We introduce new operators to the algorithm, including territories delimitation and clan migration using clustering. To validate our work, we created a dataset of more than 1.4 million tweets, on which we carried out extensive experiments. The outcomes reveal the ability of our approach to find relevant information in an effective and efficient way. They also highlight the advantages of the improved version of EHO over the original algorithm regarding different aspects. Furthermore, we undertook a comparative study with two other metaheuristic-based information foraging approaches, namely ant colony system and particle swarm optimization. Overall, the results are very promising.
\end{abstract}


\section{Introduction}
Nowadays social media are increasingly being used as an information source and people are becoming more dependent on them in their daily life. They use them to access and share information, which highly contributes to the growth of the volume of online public data. 
According to the \textit{Digital 2021 Report}, the number of social media users has increased by an average of more than 1.4 million users each day during 2020, which amounts to more than half a billion new users in 12 months \cite{WeAreSocial}. This rapid growth has propelled the total number of active social media users to 4.33 billion by April 2021, which equates to 55\% of the world’s total population.
In fact, social media are being used to seek information about serious topics, such as circulating up-to-the minute information about the Covid-19 pandemic \cite{Laato}. More generally, these platforms are frequently used by people seeking health information.
In the U.S. for instance, 80\% of Internet users search for health information online, and 74\% of them use social media \cite{Ventola}. 
In view of this impressive growth in terms of popularity and data volume of social media, the development of new large-scale information access techniques adapted to such platforms is required. 

Generally, a person engaged in an information seeking process has one or more goals in mind and uses information access tools to achieve them. Those goals can range quite widely, from finding a specific product to keeping informed about a certain topic.
Information foraging is a paradigm related to accessing information online. Usually, when people need an information, they have the opportunity to use the Web to query it. Of course information
retrieval helps to get a part of the information, thanks to the existing search engines. However, information foraging is more than just querying a search tool and getting a fragment of information. It consists in exploring the Web while using certain bio-inspired navigation mechanisms as well as Web structure related features. 
The task of foraging is grounded on the optimal foraging theory (OFT) \cite{Werner}, which paved the way to the information foraging theory (IFT) \cite{pirolli:if}. The authors of the latter studied the optimal foraging theory to understand how human users search for information. The IFT is based on the assumption that, when searching for information, humans use built-in foraging mechanisms that evolved to help our animal ancestors find food.

Recently, a significant amount of work has been done in the information access field using the \textit{information foraging theory}. Technologies and approaches such as deep learning \cite{Niu:Fan}, game theory \cite{Drias:Kechid:19}, bio-inspired computing \cite{Drias:Kechid:Pasi}, ontologies \cite{Nguyen} and multi-agent systems \cite{Drias:Kechid:17} were used for this purpose. On top of that, the IFT was exploited to solve many problems like cyber-attack prediction \cite{Dalton}, query auto-completion \cite{Jaiswal} and recommender systems \cite{Schnabel}. Some newer studies also focused on applying the IFT on social media \cite{Drias:Pasi}. \\
The aim of this article is to propose a novel bio-inspired approach to large scale information access on social media. Our approach is  based on a combination of the information foraging theory and a new enhanced elephant herding optimization that we developed for large scale information access. 
The main new contributions of the present work can be summarized as follows:
\begin{itemize}
	\item a detailed formal model for information foraging on social media;
	\item a new enhanced version of elephant herding optimization with new operators adapted to large scale information access;
	\item the use of k-means clustering to implement new operators in the enhanced elephant herding optimization;
	\item a performance evaluation on a dataset of more than 1.4 million tweets;
	\item a comparative study with other mataheuristic-based information access approaches.
\end{itemize}

The rest of this paper is organized as follows. In Section \ref{sec2}, we discuss related literature that covers some recent works on information foraging and elephant herding optimization. In section \ref{sec3}, we present our information foraging model that focuses on social media, while explaining the analogy between animal food foraging and information foraging. Section \ref{sec4} is dedicated to present our adaptation of the elephant herding optimization algorithm to information foraging on social media. In Section \ref{sec5}, we explain how we incorporate the territories concept into EHO using clustering. The Enhanced EHO for large scale information foraging is presented in Section \ref{sec6} and the experimental results are detailed in Section \ref{sec7}. Finally, we conclude in Section \ref{sec8} and discuss some future directions.

\section{Related works}\label{sec2}
This section provides a literature review in two parts.The first one reports the most recent studies on information foraging, while the second summarizes some important efforts on elephant herding optimization.

\subsection{Information foraging}
Recently, a number of studies applied the information foraging theory to address issues related to some information access approaches. 
The authors in \cite{Schnabel} explore how changes to the user interface can impact the learning accuracy of recommender systems. They use the information foraging theory to study how feedback quality and quantity are influenced by interface design choices along two axes: information scent and information access cost. To undertake a user study, the authors considered the task of picking a movie to watch. The results obtained from the use of the information foraging theory concepts such as the information scent show that the user interface factors can effectively shape and improve the implicit feedback data that is generated while maintaining the user experience. \\
In \cite{Azzopardi}, the authors measure the utility and cost of Web search engine result pages using a new measure based on the information foraging theory. According to the authors, the latter provides a number of new dimensions in which to investigate and evaluate user behavior and performance. The analysis of over 1000 popular queries issued to a major search engine show that the proposed foraging based measure provides a more accurate reflection of the utility and of observed behaviors.\\
The IFT was also used to develop standalone information access systems. 
In \cite{Drias:Kechid:19} the authors implemented a multi-agent system composed of several self-interested agents with different behaviors. The task of finding relevant information based on an information need introduced by the user was assigned to each agent. The developed system was tested on the Citation Network, which contains scientific publications along with their respective citations. The authors conducted a preprocessing step consisting in classifying the publications using the 2012 ACM ontology. The outcomes of this study demonstrate that introducing such preprocessing step in information foraging can highly contribute in making the system scalable. \\ 
Bio-inspired metaheuristics were also exploited in this context. In \cite{Drias:Kechid:Pasi:16}, the authors propose a framework for medical Web information foraging using hybrid ant colony optimization and tabu search. The experimental results on \textit{MedlinePlus} website show that the system is able to locate relevant Web pages related to specific pathologies and diseases thanks to the collaboration and self-organization aspects that characterize ant colony optimization and bio-inspired metaheuristics in general. 

\subsection{Elephant herding optimization}
The remarkable growth of the size and complexity of optimization problems made the traditional exact algorithms ineffective for solving this kind of problems \cite{Santucci}. Metaheuristic algorithms have proved to be a viable solution to this challenge. These robust algorithms, which are in most cases bio-inspired, are mainly applied to solve NP-hard problems \cite{Alirezaa, Anuar}.
Elephant Herding Optimization (EHO) is a bio-inspired metaheuristic that takes its origins from the herding behavior of elephants in nature. It was first introduced in \cite{Wang} to solve hard continuous optimization problems and has since been used to address numerous problems such as numerical optimization problems \cite{Wei}, task scheduling \cite{Sahoo}, data clustering \cite{Tuba19} and smart grid domain for Home Energy Management \cite{Mohsin}. \\
Several new EHO variants have been proposed with different improvements.
In \cite{LiGuo}, the authors introduce six individual updating strategies into basic EHO. In each strategy, one, two, or three individuals are selected from the previous iterations, and their useful
information is incorporated into the algorithm updating process. The experimental results on different test functions indicate that the proposed improved EHO version significantly outperformed basic EHO.\\
A new EHO algorithm with chaos theory to solve unconstrained global optimization problems was introduced in \cite{Tuba}. Two chaotic maps are incorporated into the basic EHO algorithm in order to improve the search quality. The comparison results with standard benchmark functions show that the new proposed algorithm outperforms the basic EHO and PSO in almost all cases.

Except for a very limited number of studies, accessing relevant information on social media based on the information foraging theory has not been addressed in the existing literature. Combining the IFT with an enhanced variant of EHO can substantially contribute in addressing the problem of large scale information access on social media.

\section{Analogy between animal food foraging and information foraging on social media}\label{sec3}
\begin{figure}[!hbtp]
	\centering
	\includegraphics[scale=0.35]{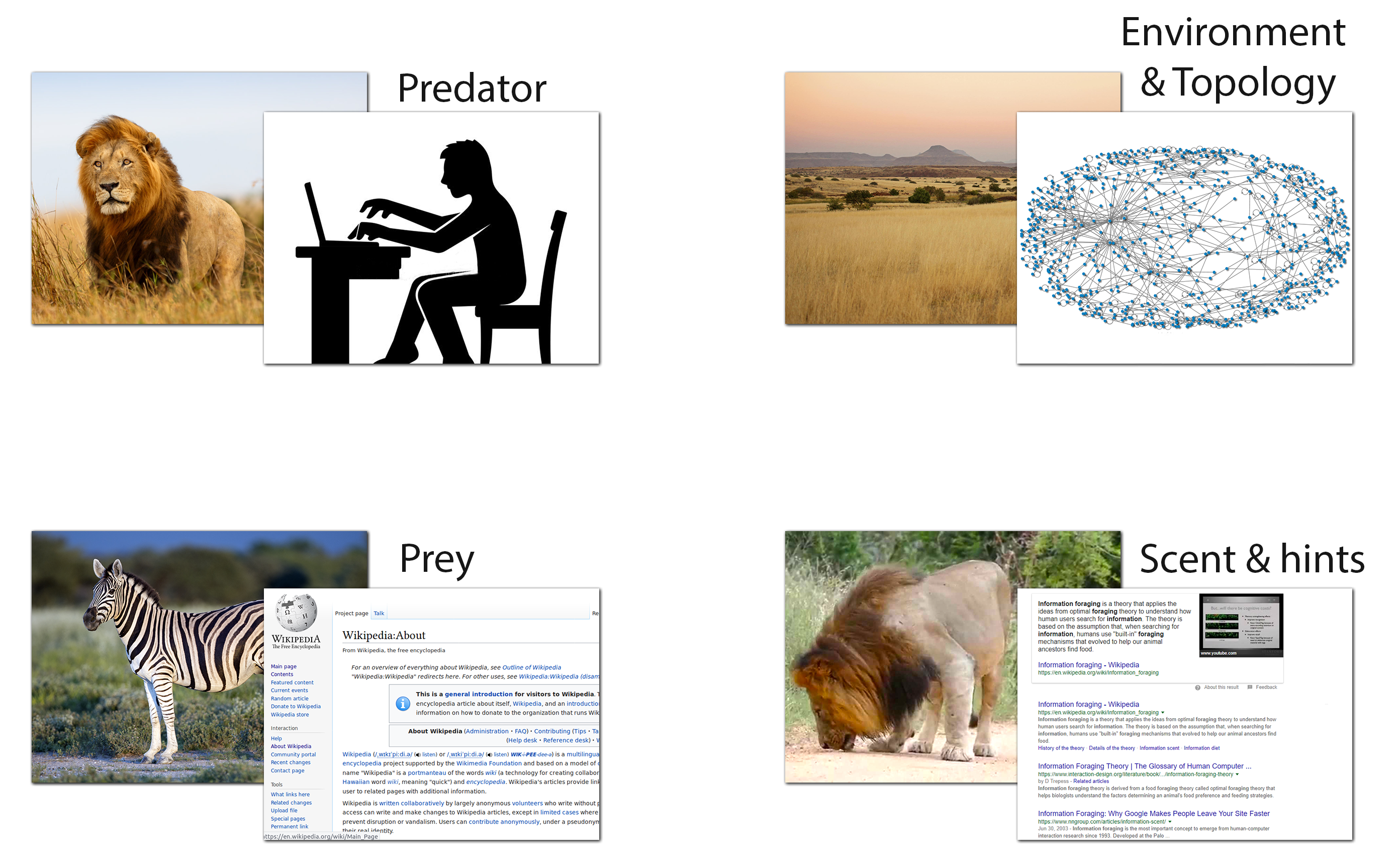}
	\caption{Analogy between Information Foraging and Food Foraging}
	\label{fig:Intro IF vs FF}
\end{figure}

The information foraging process intends to find paths leading to relevant information on the Web.  
The theory behind it is based on the analogy between animal food foraging behavior and human information seeking behavior. It assumes that when searching for information online, users follow indications and hints that guide them to relevant information, similar to how animals follow the scent of their preys to catch them.
Figure \ref{fig:Intro IF vs FF} and Table \ref{tab: IF vs FF} present a good illustration of the analogy between information foraging and animals' food foraging.

\begin{table}[!hbtp]
	\centering
	\begin{tabular}{|c|c|c|}
		\hline 
		\textbf{Elements} & \textbf{Food Foraging} & \textbf{Information Foraging} \\ 
		\hline 
		\multirow{2}{*}{Actors} & Predator & User \\ 
		\cline{2-3}
		& Prey & Relevant information \\ 
		\hline 
		\textit{Trigger} & Hunger & Information needs \\ 
		\hline 
		\textit{Environment} & Nature, wilderness & Web structure, social graph \\ 
		\hline 
		\textit{Cues} & Scent of the prey & Hyperlinks, icons, titles \\ 
		\hline 
	\end{tabular} 
	\caption{Food Foraging analogy with Information Foraging}
	\label{tab: IF vs FF}
\end{table}

The following subsections describe our proposed model for adapting information foraging theory to social media platforms, as well as the basic notions on which the analogy with animal food foraging is grounded.

\subsection{Territory:  social graph}
\begin{figure}[!htbp]
	\centering
	\includegraphics[scale=0.6]{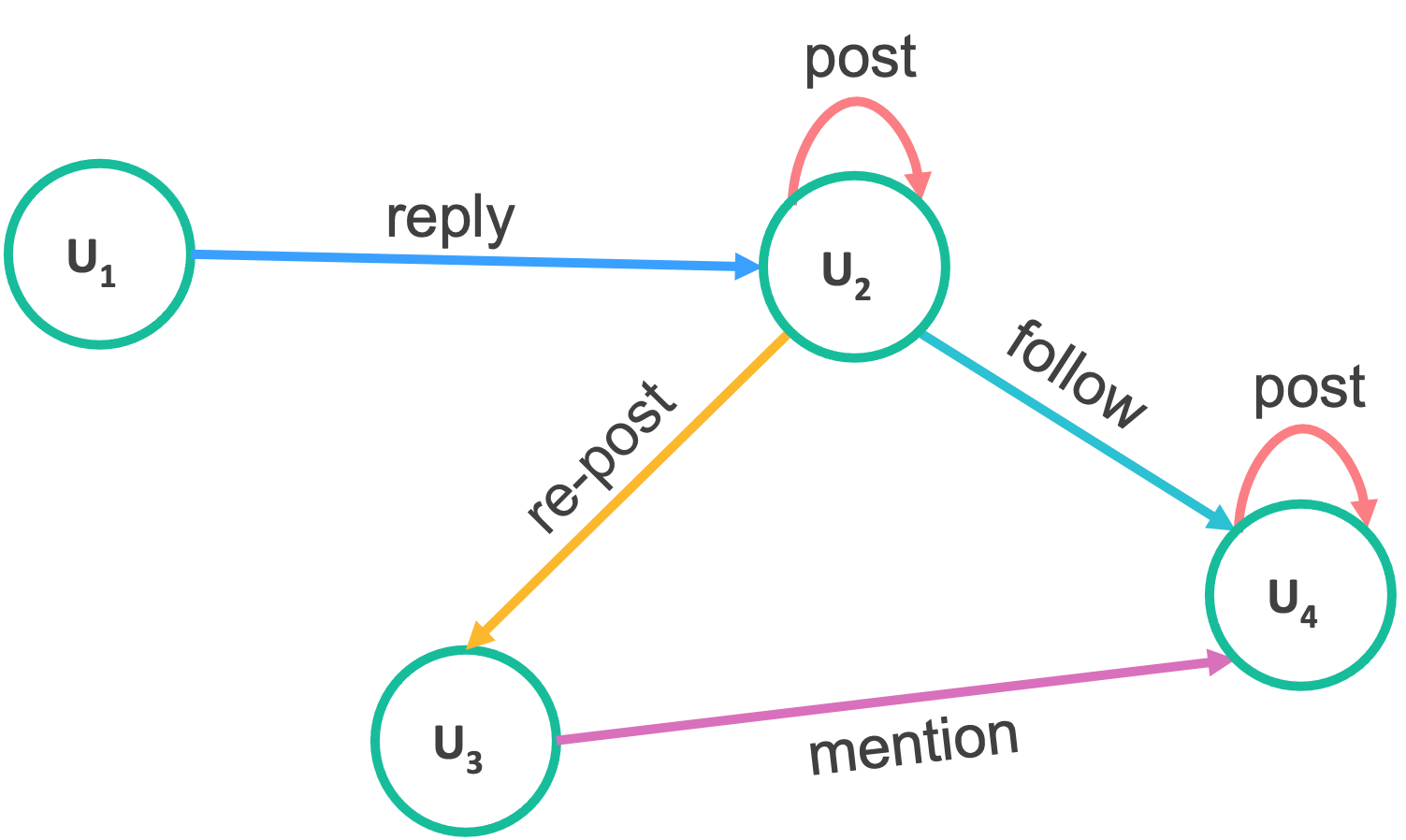}
	\caption{Social graph structure}
	\label{fig1}
\end{figure}

In the OFT, it is assumed that each animal operates in a delimited geographical territory, within which it searches for food. In information foraging, the territory corresponds to the search space composed of information sources such as documents, images and Web pages. When it comes to social media, the users' shared content serve as information sources. A social graph \cite{FacebookF8} is a representation of the users, their shared posts, and their social interactions and relationships. \\
In this paper, we model a social network as an oriented graph $G(V, E)$, where :
\begin{itemize}
	\item the set of vertices $V$ represents the social media users,
	\item the set of directed edges $E$ represents relationships and interactions in the network, such as : a post, a re-post, a friendship, a mention, a reply and a follow.
\end{itemize}
A simplified social graph structure is shown in Figure \ref{fig1}. 
The edges that reflect the relations post, re-post, mention, and reply contain the social posts and so represent the information sources. We denote the set of these \textit{content-sharing edges} by $\tilde{E}$ with $\tilde{E} \subseteq E$.

\subsection{Food diet: user's interests}
Each animal in the food chain has its own preferences in terms of food. Wilde animal for instance choose their preys based on their environment, their size and their hunting skills. In information foraging, the animal food diet is translated by the user's information needs that we call the \textit{user's interests}.

The information foraging process takes two inputs: a collection of posts represented by a social graph, in addition to the user's thematic interests. These users' interests can be expressed explicitly by the user or inferred implicitly from the user's social media activity (profile and interactions)  \cite{Drias:Pasi}.
The modeling of the user's interests consists in extracting the terms that are the most representative of the user's information needs from the keywords given by the user and/or the information accessible on their profile (biography, previous posts, etc.).

\begin{figure}[!htbp]
	\centering
	\includegraphics[scale=0.19]{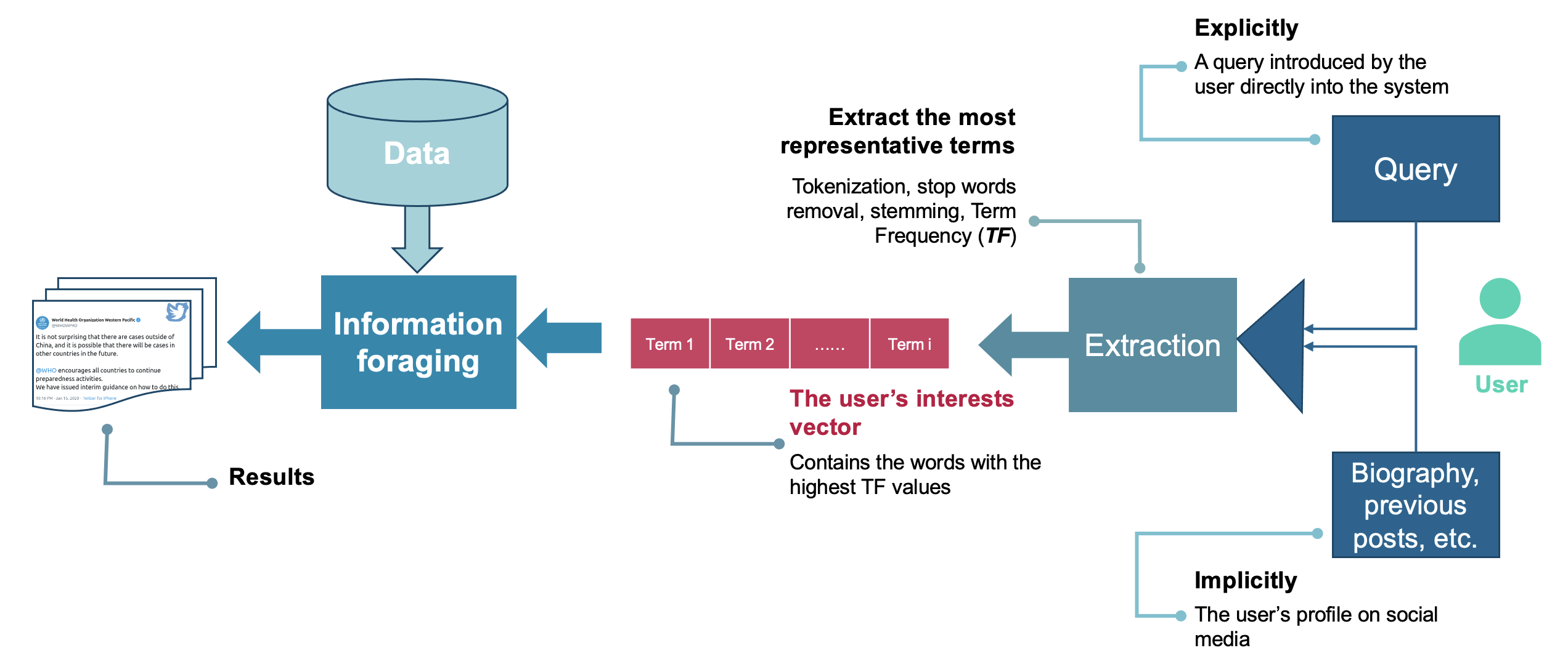}
	\caption{User's interests extraction}
	\label{user's interests}
\end{figure}

The extraction process includes: tokenization, stop words removal, stemming, and Term Frequency (TF) calculation. The words with the highest TF values are then stored into the user's interests vector $I$ following the bag-of-words model. Figure \ref{user's interests} illustrates the process of modeling the user's topical interests by a vector of terms and using it in information foraging. 

\subsection{Scent: information scent }
The general goal of information access approaches is to offer mechanisms that can help finding relevant information, while minimizing the time spent doing the search. Likewise, the goal of animals in the wild is to find a decent amount of food whilst spending less energy. To achieve that, animals generally rely on their senses to locate and hunt their preys in an effective way. The authors of the IFT notice that users have a similar behavior when looking for information on the Web. The authors assume that when browsing the Web, users exploit available hints and cues to estimate the information value contained in accessible pages and therefore decide which pages to visit. 
This can be achieved thanks to the information scent concept \cite{Budiu:Royer:Pirolli}, which can be seen in real life as the user's estimation of the value that a source of information will deliver to them. This value is primarily computed based on the source's description/content. In the case of the Web, for instance, information sources are Web pages, which are described by a URL, a title, and in certain cases an icon.

The goal in our context is to find relevant posts to satisfy a specific user's information needs. We presume that if a post is related to the user's interests, it will be more appealing to them, and that the information scent value should increase as we get closer to a relevant post and decrease otherwise. 
We define the information scent as the similarity evolution between the present post being visited at time $t$ with the user's interests vector and the considered post to be accessed in time $t+1$ with the user's interests vector. Formula (\ref{eq1}) shows how the information scent is calculated when considering to move from the current post located on the edge $\tilde{e}_i$ to one of its neighbors located on the edge $\tilde{e}_j$. 
\begin{equation}
	InfoScent(\tilde{e}_j) = Sim(\tilde{e}_j, I) - Sim(\tilde{e}_i, I) 
	\label{eq1}	
\end{equation}

Where :
\begin{itemize}
	\item $I$ is the user's interests vector;
	\item $\tilde{e}_i$ is the current post;
	\item $\tilde{e}_j \in N_i$, with $N_i$ being the set of adjacent edges to $\tilde{e}_i$, i.e. $\tilde{e}_i$'s neighborhood;
	\item $Sim()$ represents the cosine similarity between two vectors.
\end{itemize}
The main role of the information scent is to guide the foraging, a positive value indicates that we are approaching a relevant post in the social graph, whereas a negative value indicates the opposite.  

\subsection{Trail: surfing path}
While foraging food, animals follow a certain path that allows them to reach food sources in an optimal way according to the OFT. Web users have a similar behavior as they visit Web pages one at a time until reaching a relevant page that satisfies their information needs, constructing there a surfing path composed of one or more Web pages.

The information foraging process starts from an initial post and progresses through each step, attempting to reach a post with more relevant information than its predecessor. 
A surfing path is built for this purpose, starting with an initial content-sharing edge and then being enriched by adding further edges to create a chain of related posts. This means that at each step of the foraging process, the system should choose one content-sharing edge to visit from the neighborhood of the current post. This choice is made based on Formula (\ref{eq2}).

\begin{equation}
	P(\tilde{e}_i, \tilde{e}_j) = 
	\begin{cases}
		0,& \text{if } InfoScent(\tilde{e}_j) \leq 0 \\
		\frac{ InfoScent(\tilde{e}_j)}   {\sum_{\tilde{e}_l\in Np_{i}} InfoScent(\tilde{e}_l)}, & \text{otherwise}
	\end{cases}
	\label{eq2}
\end{equation}

where :
\begin{itemize}
	\item $P(\tilde{e}_i, \tilde{e}_j)$ is the likelihood of selecting the edge $\tilde{e}_j$ among the reachable edges from the current edge $\tilde{e}_i$
	\item $Np_i$ is the set of adjacent content-sharing edges of the edge $\tilde{e}_i$ with a positive information scent value, i.e. $\forall \tilde{e}_l\in Np_{i}$ $InfoScent(\tilde{e}_l) > 0$. 
\end{itemize}

\section{Adapted elephant herding optimization to information foraging}\label{sec4}
\begin{figure}[!htbp]
	\centering
	\includegraphics[scale=0.55]{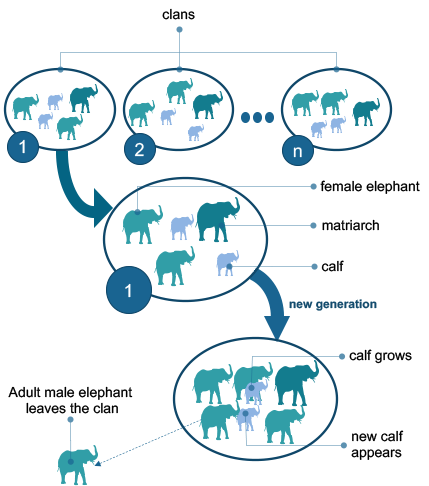}
	\caption{Elephants society structure}
	\label{fig:EHO}
\end{figure}

The social structure of elephants is complex, varying by gender and population dynamics. Adult females form a matriarchal societies, while adult males are usually solitary \cite{Elephants:Charif, Elephants:Fritz}. 
A herd structure is similar to concentric rings, with the innermost circle comprising a family unit of related female adults. A family unit is formed by the eldest most dominant female called the matriarch as well as her adult daughters, their calves and a number of juveniles. The male calves leave the herd when reaching adulthood, generally between the age of 12 and 15. From this stable core, the groupings widen to include less familiar individuals. A clan is formed when elephants gather in large groups consisting of different herds.
The functioning of the elephants society is illustrated by Figure \ref{fig:EHO}.

We decided to combine the information foraging theory with EHO in order to efficiently identify useful information in large social graphs. Each elephant will look for relevant social posts by browsing a section of the graph in this manner, constructing surfing paths that lead to relevant information. The elephants perform the foraging while taking advantage of the hierarchy and organization of their society. This will allow them to collaborate and find relevant information in a more effective and efficient way. \\
In this section, we adapt the basic Elephant Herding Optimization algorithm to information foraging.
Note that the original EHO was developed to address continuous problems, whereas information foraging is a discrete combinatorial problem. Further improvements to basic EHO will be introduced in sections \ref{sec5} and \ref{sec6}.

\subsection{Generating the elephant population and assigning the positions} 
\begin{figure}[!htbp]
	\centering
	\includegraphics[scale=0.33]{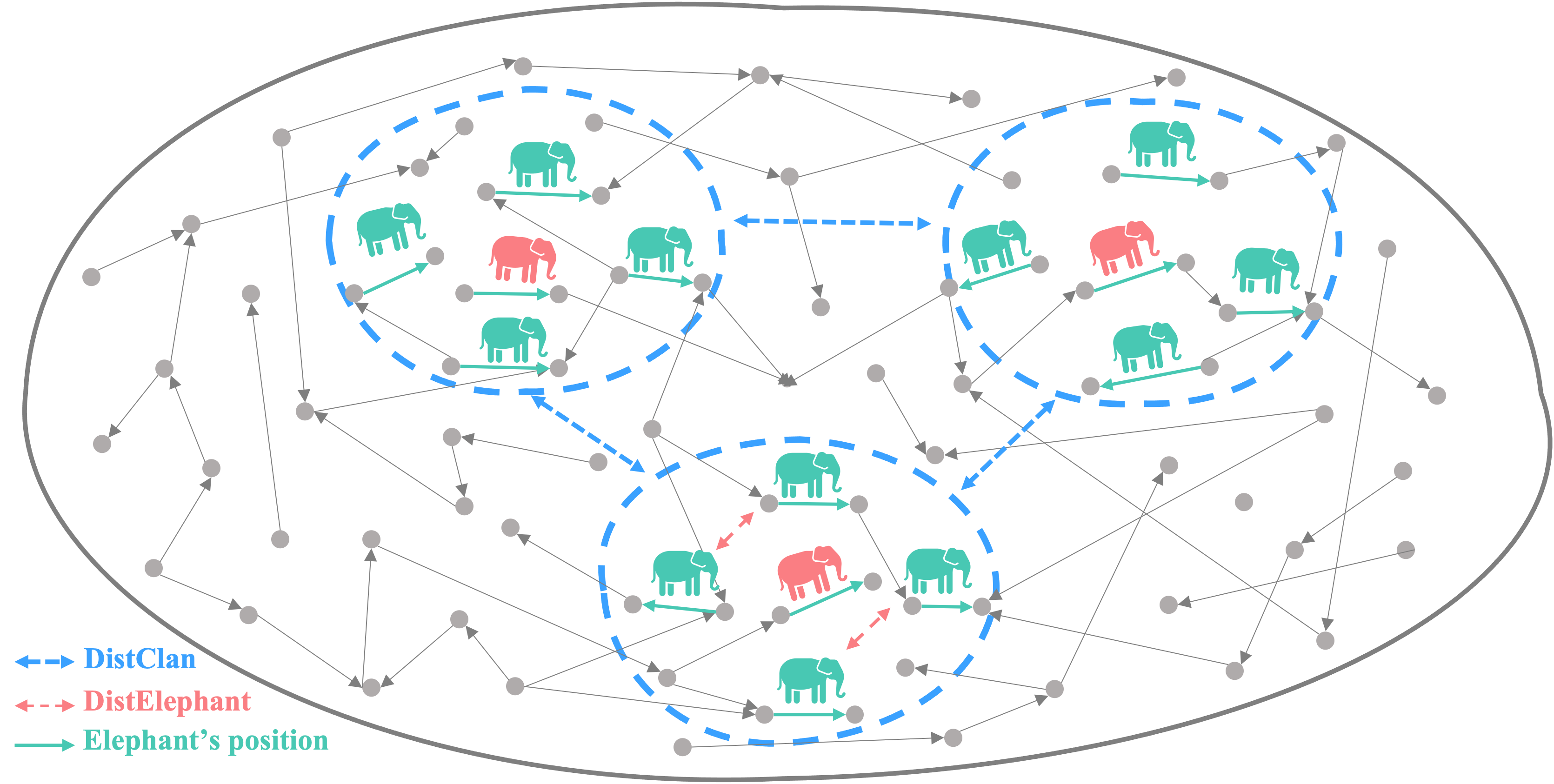}
	\caption{A population of elephants distributed over a social graph}
	\label{fig2}
\end{figure}

To create an elephant population with $p$ clans, we first generate $p$ different elephants with respect to \textit{distClan}, which represents the shortest minimal distance between clans. The rest of the elephants in each clan are then formed using the positions of the $p$ elephants, with respect to \textit{distElephant}, which represents the maximum distance between elephants in the same clan. 
Figure \ref{fig2} depicts a population of three clans scattered throughout a social graph.

We assign $m$ different positions to a social graph with $m$ content-sharing edges, one for each edge. These positions are represented by integer values in the interval $[1, m]$. Each elephant $j$ belonging to clan $c_i$ is identified by a unique position denoted by the $x_{c_i,j}$. An elephant's position at time $t$ is the position of the edge it is visiting at that time.

\subsection{Surfing paths construction}
Each elephant is assigned the duty of building a surfing path leading to relevant information during one iteration of the EHO algorithm. This is accomplished using Algorithm (\ref{alg1}).

\begin{algorithm}
	\renewcommand{\algorithmicrequire}{\textbf{Input:}}
	\renewcommand{\algorithmicensure}{\textbf{Output:}}
	\begin{algorithmic}[1]
		\Require $x_{c_i,j}$: elephant's position, $I$: user's interests, $G$: social graph;
		\Ensure $SP$: a surfing path leading to a relevant post; 
		\State $SP \leftarrow \emptyset$.
		\State Locate the edge $\tilde{e}_i$ corresponding to the elephant's initial position $x_{c_i,j}$
		\State $SP = SP \cup \lbrace \tilde{e}_i \rbrace $  \label{loop1}
		\State $N_i \leftarrow \emptyset$
		\ForAll {adjacent edge $\tilde{e}_j$ to $\tilde{e}_i$ }
		\State Calculate $InfoScent(\tilde{e}_j)$ using Formula (\ref{eq1})
		\If {$InfoScent(\tilde{e}_j) > 0$} $N_i = N_i \cup \lbrace \tilde{e}_j \rbrace $
		\EndIf
		\EndFor
		\If {$N_i = \emptyset $} return $SP$
		\Else \State Select a new content-sharing edge to visit from $N_i$ following Formula (\ref{eq2})
		\State Go to \ref{loop1}	
		\EndIf
	\end{algorithmic}
	\caption{Building a surfing path}
	\label{alg1}
\end{algorithm}

\subsection{Solutions evaluation}
At the end of each iteration of the algorithm, the solutions (the surfing paths) fetched by the elephants are evaluated according to a fitness function. To do so, we compute the similarity between the user's interests and each solution using Formula (\ref{eq3}).

\begin{equation}
	f(x_{c_i,j}) = Sim(\tilde{e}_k, I)
	\label{eq3}	
\end{equation}

Where :
\begin{itemize}
	\item $\tilde{e}_k$ represents the last social post on the surfing path constructed by elephant $j$ in clan $c_i$;
	\item $I$ is the user's interests vector;
	\item $Sim(\tilde{e}_k, I)$ represents the cosine similarity between $\tilde{e}_k$ and $I$.
\end{itemize}

\subsection{Updating Operator}
The elephants' positions are updated using Formula \eqref{eq4} at the end of each iteration of the algorithm, once the new solutions have been evaluated.

\begin{equation}
	x_{new,c_i,j} = x_{c_i,j} + \alpha (x_{best,c_i} - x_{c_i,j}) \times r
	\label{eq4}	
\end{equation}

Where: 
\begin{itemize}
	\item $x_{new,c_i,j}$ : is the new position of the elephant;
	\item $x_{c_i,j}$  : is the current position of the elephant;
	\item $x_{best,c_i}$ : is the matriarch's position;
	\item $ \alpha \in [0,1]$ : is an empirical parameter that defines the influence of the matriarch over the new position of the elephant $j$;
	\item $r \in [0,1]$ : is a random number, which aims at improving the diversity.
\end{itemize}

The position of each clan's matriarch is also updated throughout generations by utilizing Formula \eqref{eq5} to calculate the average fitness of each clan. After that, Formula \eqref{eq6} is used to compute the new matriarch's position using the position of the elephant with the closest fitness value to the clan's average fitness. 

\begin{equation}
	f_{avg, c_{i}} = \frac{1}{n_{c_i}} \sum_{j=1}^{n_{c_i}} f(x_{c_i,j}) 
	\label{eq5}	
\end{equation}

\begin{equation}
	x_{new best,c_i} = x_{avg,c_i} \times \beta                                    
	\label{eq6}	
\end{equation}

Where:
\begin{itemize}
	\item $f_{avg,c_i}$: represents the average fitness value of the clan $c_i$;
	\item $x_{new best,c_i}$: represents the new position of the matriarch of the clan $c_i$;
	\item $x_{avg,c_i}$: is the position of the elephant with the closest fitness value to $f_{avg,c_i}$;
	\item $\beta \in [0,1]$: is an empirical parameter, which determines the influence of the average position on the matriarch's new position;
	\item $n_{c_i}$: represents the number of elephants in the clan $c_i$;
	\item $x_{c_i,j}$ : is the position of elephant $j$ in clan $c_i$.
\end{itemize}

\subsection{Separating Operator }
The elephant with the lowest fitness value will leave the clan at the end of each generation. Formula \eqref{eq7} is used to create a new elephant to replace the one that left.

\begin{equation}
	x_{worst} = x_{min} + (x_{max} - x_{min} + 1) \times r
	\label{eq7}
\end{equation}	

Where:
\begin{itemize}
	\item $x_{worst}$ stands for the position of the elephant with worst fitness value;
	\item $x_{min}$  and $x_{max}$ are the upper and lower bounds of the positions interval;
	\item $r $ is a stochastic and uniform distribution parameter.
\end{itemize}	

\section{Defining territories with clustering}\label{sec5}
Although elephants are not territorial animals, they utilize specific home areas during particular times of the year. Their home ranges vary from from 15 to 3,700 square kilometers (24 to 5,958 square miles) depending on the population and the habitat. This delimited area helps elephants to better master their environment and remember the location of food and water sources \cite{IUCN}.\\
Implementing this concept and incorporating it into EHO would be of a great benefit to solve large scale problems. In fact, dividing the search space into sub-areas based on some problem-related features can help to limit the search to one of these sub-areas, and thus improve the effectiveness and efficiency of EHO.    

When constructing a surfing path during the information foraging process, the choice of the starting point can have a major impact on the outcome and therefore, determine whether or not the path will lead to relevant information. In fact, locating the right post from which to initiate the navigation with regards to the user's interests is of a high importance. This becomes even more obvious and crucial when dealing with large scale data, as skipping posts that are not related to the user's interests could significantly improve the efficiency of the information foraging process. For instance, if a user is interested in information about health, it would be unnecessary to search posts talking about sport.\\
Consequently and given the above observations, we propose to introduce a new step in the Elephant Herding Optimization algorithm, consisting of dividing the search space into multiple regions in order to explore them more efficiently. In addition to modeling the concept of territories in nature, this will improve the performance of the algorithm, especially when dealing with large scale problems. \\
There are numerous methods for grouping similar objects together, they can be either supervised or unsupervised depending on whether classes of objects already exist or not. Supervised classification considers classes to insert the objects whereas unsupervised classification generates clusters as outcome.
Clustering is the process of organizing objects into groups whose members are similar. A cluster is a collection of objects which are consistent internally, but clearly dissimilar to the objects belonging to other clusters. One of the main advantages of clustering over supervised classification is the fact that it doesn't require predefined classes and It can therefore be performed with data of different sizes without the need of a taxonomy or a training set. 

In this paper, we perform the clustering phase using k-means algorithm, which is known for its efficiency with large datasets and its capacity of working with textual data \cite{WangZhou, Dobsa}. It intends to automatically group a set of $n$ objects into $k$ clusters, so that objects in a same cluster are similar to one another while objects from different clusters are dissimilar \cite{MacQueen}. The grouping decision is based on the distance between the object and each cluster centroid (mean), in a way that each object belongs to the cluster with the nearest centroid. A centeroid serves as a prototype of its corresponding cluster, and is defined as the average of all the objects in that cluster. The number of clusters $k$ can be either predefined or user-defined depending on the problem, the number of objects, and the goal behind clustering. \\
When it comes to textual data clustering, the idea consists in grouping texts or documents in clusters based on their content similarity. In order to achieve this goal, a proper document representation method is necessary. We use the vector space model (VSM) to represent each social post $\tilde{e}_i$ as a weighted vector of terms $\tilde{e}_i=<w_{i1}, w_{i2}, ..., w_{i |T|}>$ 
where $T$ is the set of terms or features that occur at least once in the social graph $G$. 
The detailed clustering process using k-means is presented in the following subsections. 

\subsection{Text preprocessing}
This preprocessing phase refers to the set of actions that are applied to the social posts in order to achieve a good statistical representation of the whole collection. This phase is performed before defining the weights of the words using TF-IDF and includes the following actions:
\begin{itemize}
	\item Tokenization, which consists in dividing the text into individual words.
	\item Removing special characters related to social media platforms, links, usernames, etc. 
	\item Deleting common words that don't bring any semantic meaning to the text using a stop words list.
	\item Reducing each word to its root using adequate algorithms such as \textit{Porter Stemming}. As a result, inflected words will be grouped under their word stem, which is referred to as a term. 
\end{itemize}

The result of the textual preprocessing is a collection of posts that are each represented by a set of significant terms. The following subsection explain how each post is afterwards converted into a weighted vector of terms. 

\subsection{Feature extraction with TF-IDF}
The goal of the feature extraction using TF-IDF is to create a mapping of the textual data into vectors of terms. This vector representation of the social posts is grounded on the term relevance concept. The weight associated to each term should be proportional to its importance, so that terms with high weight values are considered as relevant. This method consists in increasing the weight of a term when it appears many times in a post and lowering it when it is common in many posts. We can summarize the TF-IDF  calculation with the two following steps:

\subsubsection{Term Frequency (TF)}
The term frequency  $TF(t_i, \tilde{e}_j)$ estimates the importance of a term $t_i$ in a post $\tilde{e}_j$ based on how often $t_i$ appears in $\tilde{e}_j$. The more frequent a term is in a post, the more important it is in its description.
We use Formula (\ref{eq:tf}) to compute the term frequency. 

\begin{equation}
	TF(t_i, \tilde{e}_j) = \dfrac{freq_{ij}}{\sum_{\forall t_l \in \tilde{e}_j} freq_{lj}}
	\label{eq:tf}
\end{equation}

With $freq_{ij}$ being the number of occurrences of term $t_i$ in post $\tilde{e}_j$.

\subsubsection{Inverse Document Frequency (IDF)}
Inverse Document Frequency indicates how commonly a word is used in a collection of documents. A term has an important characterizing power if its frequency is high in a particular social post and low in the rest of the posts. We estimate the inverse document frequency using Formula \eqref{eq:idf}. 

\begin{equation}
	IDF(t_i) = log(\dfrac{\arrowvert\tilde{E}\arrowvert}{n_i})
	\label{eq:idf}
\end{equation}

Where: 
\begin{itemize}
	\item $\arrowvert\tilde{E}\arrowvert$ is the total number of posts in the social graph. 
	\item $n_i$ is the number of posts containing the term $t_i$.
\end{itemize}

\subsubsection{TF-IDF}
We compute the weigh of a term $t_i$ in a post $\tilde{e}_j$ using the $TF-ID$ score, which is obtained by Formula \eqref{eq:tf-idf}. 
The value of the weight increases proportionally to the number of times a term appears in the post and is offset by the number of posts containing that term.
\begin{equation}
	w_{ij} = TF-IDF(t_i, \tilde{e}_j) = TF(t_i, \tilde{e}_j) \times IDF(t_i)
	\label{eq:tf-idf}
\end{equation}

Table \ref{tab:tf-idf} illustrates the vector representation of a social graph containing $p$ posts and $n$ terms using the vector space model with TF-IDF weighting measure. 

\begin{table}[!hbtp]
	\centering
	\begin{tabular}{|c|c|c|c|c|c|}
		\hline
		\backslashbox{Terms}{Posts}	& \hfil $\tilde{e}_1$ & \hfil $\tilde{e}_2$ & \hfil $\tilde{e}_3$ & \hfil ... & \hfil $\tilde{e}_p$ \\
		\hline
		\hfil $term_1$	& \hfil 0.015  & \hfil 0.342 & \hfil 0 & \hfil ... & \hfil 0 \\
		\hline
		\hfil $term_2$	& \hfil 0.231 & \hfil 1.164 & \hfil 0.324 & \hfil ... & \hfil 1.002 \\
		\hline
		\hfil $term_3$	& \hfil 0 & \hfil 0.102  & \hfil 0 & \hfil ... & \hfil 0.076 \\
		\hline
		\hfil ...	& \hfil ... & \hfil ... & \hfil ... & \hfil ... & \hfil ... \\
		\hline
		\hfil $term_n$	& \hfil 1.562 & \hfil 0 & \hfil 0.067 & \hfil ... & \hfil 0 \\
		\hline
	\end{tabular}
	\caption{Vector representation with TF-IDF}
	\label{tab:tf-idf}
\end{table}

Once the posts are cleaned and represented as weighted vectors of terms, the clustering phase can be launched using k-means algorithm.  

\subsection{Clusters initialization}
A centroid is assigned to each cluster amongst the posts of the social graph. Each centroid is represented by the weighted terms vector of its corresponding post. The $k$ centroids $m_1, m_2, ..., m_k$ are initialized randomly by choosing $k$ random posts. The pseudo code of the initialization is presented in Algorithm (\ref{alg3}).

\begin{algorithm}
	\renewcommand{\algorithmicrequire}{\textbf{Input:}}
	\renewcommand{\algorithmicensure}{\textbf{Output:}}
	\begin{algorithmic}[1]
		\Require $\tilde{E}$: posts of the social graph $G$, $k$: number of clusters;
		\Ensure $k$ centroids;
		\For{$i \leftarrow 1$ to $k$}
		\State $r \leftarrow $ random position 
		\State $m_i \leftarrow \tilde{e}_r$
		\State Insert $m_i$ in $Centroids$ 
		\EndFor
		\State Return $Centroids$
	\end{algorithmic}
	\caption{Centroids Initialization}
	\label{alg3}
\end{algorithm}

Once the centroids defined, the clusters are populated with posts based on the distance between each post and the clusters centroids. 
To compute this distance, we use the Euclidian distance measure, which represents the ordinary straight-line distance between two points in Euclidean space. In our case each point is either a post $\tilde{e}_i$ or a centroid $m_j$, both represented by a weighted terms vector. In an n-dimensional space, the distance is calculated using Formula \eqref{eq:euclidianDist}.

\begin{equation}
	d(\tilde{e}_i, m_j) = \sqrt{ \sum_{l=1}^{n} {(\tilde{e}_{il} - m_{jl})}^{2} } 
	\label{eq:euclidianDist}	
\end{equation}

Where:
\begin{itemize}
	\item $n$ : represents the vector's size;
	\item $m_{jl}$: represents the weight of term $l$ in the centroid $m_j$;
	\item $\tilde{e}_{il}$: represents the weight of term $l$ in post $\tilde{e}_i$.
\end{itemize}

The pseudo code of the clusters construction is presented in Algorithm (\ref{alg4}).

\begin{algorithm}
	\renewcommand{\algorithmicrequire}{\textbf{Input:}}
	\renewcommand{\algorithmicensure}{\textbf{Output:}}
	\begin{algorithmic}[1]
		\Require $\tilde{E}$: posts of the social graph $G$, $Centroids$: centroids vectors;
		\Ensure $S$: set of clusters;
		\ForAll {$\tilde{e}_i \in \tilde{E}$}
		\State $min_{dist} = \infty$
		\State $cluster_{id} \leftarrow -1 $
		\ForAll {$m_j \in Centroids$}
		\State compute the distance $d(\tilde{e}_i, m_j)$ using Formula (\ref{eq:euclidianDist})
		\If{$ d(\tilde{e}_i, m_j) < min_{dist} $} 
		\State $cluster_{id} \leftarrow j$ 
		\EndIf
		\EndFor
		\State Insert $\tilde{e}_i$ in the right cluster $S_{cluster_{id}} $ 
		\EndFor
		\State Return $S$
	\end{algorithmic}
	\caption{Clusters construction}
	\label{alg4}
\end{algorithm}

\subsection{Centroids update}
The centroids are updated throughout the iterations of the k-means algorithm. Their positions are recalculated and moved towards the center of their respective clusters at the end of each iteration. For this purpose, the mean weighted terms vector $\mu$  is computed for each cluster. This vector represents the average weight of each term in all the posts belonging to the cluster, and will be used to define the new cluster's centroid. After generating the vector $\mu_j$ of the cluster $S_j$, the distance between this vector, which doesn't constitute a real social post, and each post of the cluster $S_j$ is calculated in order to select the nearest post to $\mu_j$ and set it as the new centroid $m_j$. This process is detailed in Algorithm (\ref{alg5}). 

\begin{algorithm}
	\renewcommand{\algorithmicrequire}{\textbf{Input:}}
	\renewcommand{\algorithmicensure}{\textbf{Output:}}
	\begin{algorithmic}[1]
		\Require $S_j$: a cluster;
		\Ensure $m_j$: the new cluster's centroid;
		\State $\mu_j \leftarrow [ 0, 0, ..., 0]$
		\ForAll {$\tilde{e}_i \in S_j$}
		\State $\mu_j \leftarrow \mu_j + \tilde{e}_i $
		\EndFor
		\State $\mu_j \leftarrow \frac{1}{\mid S_j \mid} \mu_j $
		\State $min_{dist} = \infty$
		\State $post_{id} \leftarrow -1 $
		\ForAll {$\tilde{e}_i \in S_j$}
		\State compute the distance $d(\tilde{e}_i, \mu_j)$ using Formula (\ref{eq:euclidianDist})
		\If{$ d(\tilde{e}_i, \mu_j) < min_{dist} $} 
		\State $post_{id} \leftarrow i$ 
		\EndIf
		\EndFor
		\State $m_j \leftarrow \tilde{e}_{post_{id}} $
		\State Return $m_j$
	\end{algorithmic}
	\caption{Centroids update}
	\label{alg5}
\end{algorithm}

\subsection{K-means algorithm for territories definition}
K-means algorithm is launched prior to the information foraging process, in order to define search territories and therefore divide the search space into multiple clusters based on the content of the social posts. Once the centroids are initialized, the algorithm enters a loop composed of two main steps. The first takes in charge the clusters creation by assigning each post of the social graph to the cluster whose centroid is the nearest. The second step defines a new centroid for each cluster, based on the mean of the weighted terms vectors of all posts assigned to that cluster. The stop condition of the loop is either convergence or the reach of a maximum number of iterations. 
The territories definition process using k-means clustering is presented in Algorithm (\ref{alg6}). 

\begin{algorithm}
	\renewcommand{\algorithmicrequire}{\textbf{Input:}}
	\renewcommand{\algorithmicensure}{\textbf{Output:}}
	\begin{algorithmic}[1]
		\Require $\tilde{E}$: posts of the social graph $G$;
		\Ensure $S$: set of $k$ clusters;
		\State initialize $k$ centroids randomly using Algorithm (\ref{alg3})
		\Repeat 
		\State create $k$ clusters using Algorithm (\ref{alg4})
		\State update the centroids using Algorithm (\ref{alg5})
		\Until{convergence or max\_itererations}
		\State return S
	\end{algorithmic}
	\caption{K-means for territories definition}
	\label{alg6}
\end{algorithm}

Note that the time complexity of the algorithm is estimated to $\mathcal{O}(n*m*k*l)$, with $n$ being the total number of posts in the social graph, $m$ the size of the vectors, $k$ the number of clusters and $l$ the number of iterations. Although the complexity is linear, it can require a  significant time especially with large scale social graphs. Nevertheless, this will not affect the information foraging performance, since the clustering is performed  offline and only once. 

\section{Enhanced EHO for large scale information foraging (EEHOLSIF)} \label{sec6}
Addressing large scale information foraging can be tricky and time consuming. The worst case complexity of information foraging corresponds to the case when the social graph is a complete graph and is estimated to $\mathcal{O}(\prod_{i=1}^{p} n- i)$, with $n$ being the total number of posts in the graph and $p$ the surfing depth \cite{Drias:Kechid:Pasi:16}. 
In this paper, we propose a new bio-inspired approach to information access based on enhanced elephant herding optimization using the concepts presented in sections \ref{sec3}, \ref{sec4} and \ref{sec5}. We introduce a new enhanced Elephant Herding Optimization variant to improve the performance of the original algorithm and adapt it to large scale information foraging.
Our contribution focuses on several aspects, from the initialization of the algorithm to the clans' structure. The main aspects are detailed in the following subsections. 

\subsection{Semantic position assignment}
The concept of territories introduced in section \ref{sec5} allows to considerably optimize the foraging process by delimiting the search area.  We exploit the clustering results to assign a numerical position to each post on the social graph based on the posts' content. 
Following the clustering process, each post of the social graph is given an integer identifier, which will serve as a position in EEHOLSIF algorithm, in a way that posts belonging to the same cluster have neighbor positions. These positions are sorted according to the Euclidian distance between posts and the centroid of the cluster, so within the same cluster if a position $i$ is less than another position $j$, this means that the post associated with $i$ is closer to the centroid than the post associated with $j$. For instance, let us consider a social graph with 4000 posts, and 3 clusters with $S_1$ having 1500 posts, $S_2$ having 1500 posts and $S_3$ having 998 posts. The positions in each cluster will be distributed as follows: 
\begin{itemize}
	\item $S_1$= $\bigcup\limits_{i=1}^{1500} \tilde{e}_i$ 
	\item $S_2$= $\bigcup\limits_{i=1501}^{3001} \tilde{e}_i$ 
	\item $S_3$= $\bigcup\limits_{i=3002}^{4000} \tilde{e}_i$
\end{itemize}

\subsection{Initialization of the algorithm}
Territories definition and semantic positions assignment will play a major role in the initialization phase of EEHOLSIF. Indeed, unlike the original version of the algorithm, where the initialization is performed in a complete random way, the clustering and the semantic positions assignment permit to target the cluster containing the posts that are the closest to the user's interest and then set the initial elephants' positions accordingly. The minimal distance between clans \textit{distClan} and the maximal distance between elephants of the same clan \textit{distElephant} become more representative since the positions are assigned based on the content of the posts. In fact, this ensures that the elephants of the same clan are browsing pots that have similar content while elephants of different clans are located on dissimilar posts with regards to their content. This will result in a better distribution of the elephants on the search space and a better coverage of the potential solutions. 

In order to initialize the clans, we first compute the euclidean distance between the user's interests vector $I$ and each centroid vector $m_j$. Then, we launch the different clans either on the cluster with the nearest centroid or on a cluster chosen according to a uniform distribution probability. For this purpose, we introduce $q$, a random variable uniformly distributed in $[0,1]$ and $q_0 \in [0,1]$ a tunable parameter. We propose the pseudo random proportional rule for choosing a territory highlighted by Formula \eqref{eq:claninit}.

\begin{equation}
	\begin{aligned}
		\text{if } q < q_0 & \text{ then }\\ 
		& P(c_i, S_j) = 	
		\begin{cases}
			1 & \text{if } j = argmin(d(I, m_j)) \\
			0 & \text{otherwise}
		\end{cases}\\
		& \text{else } \\
		&  P(c_i, S_j) = \dfrac{d(I, m_j)}{\sum_{l = 1}^{k} d(I, m_l)}
	\end{aligned}
	\label{eq:claninit}
\end{equation}

Where: 
\begin{itemize}
	\item $P(c_i, S_j)$ is the probability to place clan $c_i$ on cluster $S_j$;
	\item $d(I, m_j)$ is the euclidean distance between the user's interests $I$ and the centroid of cluster $S_j$
\end{itemize}

\subsection{Solution construction}
Another improvement is related to the construction of the solution, which is in our case a surfing path. Unlike in the original algorithm where the elephants consider ready-made solutions, we give the task of constructing a solution to each elephant starting from its initial position. For this purpose, we incorporate information foraging concepts including the information scent (Formula (\ref{eq1})) and the surfing decision rule (Formula (\ref{eq2})). The solution construction process is given by Algorithm (\ref{alg1}). \\
Note that during the solution construction, an elephant can leave its territory if the surfing path leads it towards a post located on a neighbor territory. We consider two territories as neighbors if they share adjacent edges. Two edges are adjacent if they are both incident with a common vertex. 

\subsection{Clan migration}
In nature, an elephant clan might separate from the larger herd in response to limited food supplies encountered during a dry season. If food sources are scarce, it is more efficient for elephants to travel as individual clans, rather than large herds.\\
We incorporate this natural phenomenon in the enhanced EHO for large scale information foraging as a stagnation prevention mechanism. We believe there is a strong analogy between the lack of food sources in nature on one side and the inability of the elephants to improve their solution after several generations on the other. 
We introduce a migration parameter $t_0$, which serves as a threshold that controls the maximum number of generations a clan can spend without improving its best solution. If a clan exceeds this threshold, it migrates towards a new territory with the hope of finding better solutions. The migration is performed by choosing a new cluster randomly and defining the migrating clan positions within that cluster.

The enhanced elephant herding optimization for large scale information foraging is presented in Algorithm (\ref{alg7}).

\begin{algorithm*}
	\renewcommand{\algorithmicrequire}{\textbf{Input:}}
	\renewcommand{\algorithmicensure}{\textbf{Output:}}
	\begin{algorithmic}[1]
		\Require $I$: the user's interests, $G$: the social graph;
		\Ensure a list of surfing paths ranked by relevance; 
		\State Divide the search space into $k$ different territories using Algorithm (\ref{alg6}) 
		\State Set the generations counter $t \leftarrow 1$, the solutions list $sols \leftarrow \emptyset$ and the stagnation counter for each clan $g_{c_i} \leftarrow 0$
		\State Initialize empirical parameters $\alpha$, $\beta$, $q_0$, $t_0$, maximum generations $MaxGen$, number of clans $nClans$, population size and number of elephants in each clan $n_{c_i}$.
		\State Initialize the positions of the elephants according to the user's interests $I$ using Formula (\ref{eq:claninit}) and with respect to $distClan$ and $distElephant$
		\While {$t \leq MaxGen$}
		\For {$i \leftarrow 1$ to nClans} \Comment{for all clans in elephant population}
		\For {$j \leftarrow 1$ to $n_{c_i}$} \Comment{for all elephants in clan $c_i$}
		\State Build the elephant's $j$ surfing path using Algorithm (\ref{alg1})
		\State Calculate the elephant's fitness using Formula \eqref{eq3}.
		\EndFor
		\If{clan $c_i$ improved its best solution compared to the previous generation} 
		\State Update $bestSol_{c_i}$ 
		\Else  
		\State $g_{c_i} \leftarrow g_{c_i} + 1$ 	
		\EndIf
		\If {$g_{c_i} \geq t_0 $}
		\State Migrate clan $c_i$ towards a new territory chosen randomly
		\State  $g_{c_i} \leftarrow 0$
		\Else 		
		\State Update the positions of the elephants $x_{c_i,j}$ using Formula \eqref{eq4}.
		\State Update the matriarch's position using Formula \eqref{eq6}.
		\State Locate the worst elephant to leave clan $c_i$ according to the fitness function. 
		\State Generate a new elephant in the clan $c_i$ using Formula \eqref{eq7}.
		\EndIf
		\EndFor
		\State Append the best surfing paths found in generation $t$ to $sols$
		\State Update the generation counter, $t \leftarrow t + 1$. 
		\EndWhile
		\State Return the best surfing paths ranked relevance. 
	\end{algorithmic}
	\caption{Enhanced EHO for large scale information foraging}
	\label{alg7}
\end{algorithm*}

\section{Experiments}\label{sec7}
This section is organized in four subsections, first we describe the dataset we use in the evaluation. We then present the results obtained with the adapted EHO for information foraging. Next, we follow up with the results of the enhanced EHO for large scale information foraging. Finally, we finish by doing a comparative study with other approaches from the literature. \\  
All the experiments were implemented using \textit{Java} programming language and were conducted on a laptop running Windows 10 with an Intel Core i5-4300M CPU at 2.60GHz and 6GB of RAM. 

\subsection{Dataset description }
\begin{figure}[!hbtp]
	\centering
	\includegraphics[scale=0.55]{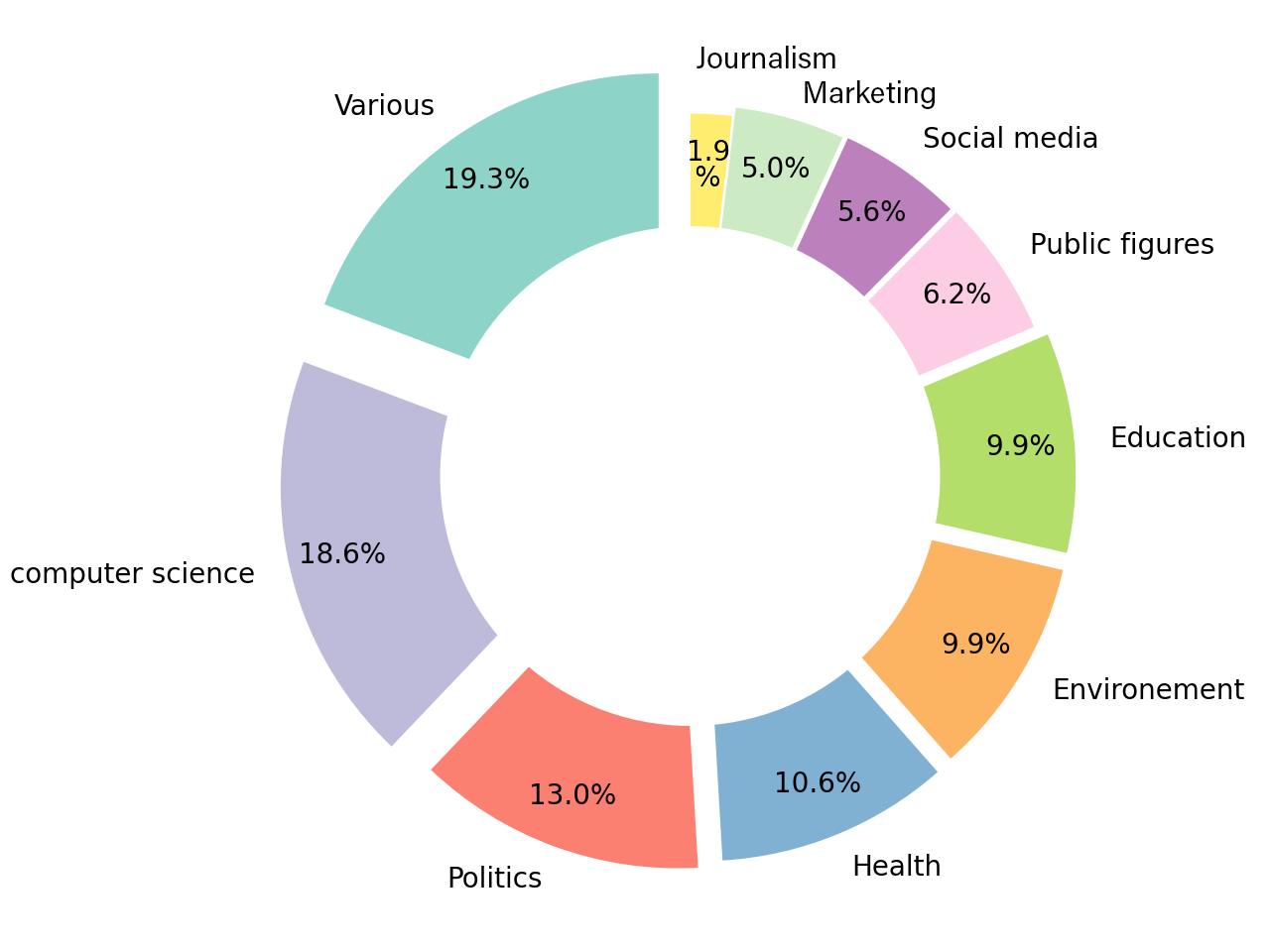}
	\caption{Topics covered by the dataset}
	\label{fig3}
\end{figure}

We tested our algorithm on \textit{Twitter}, which is one of the most popular social networks and microblogging platforms. We constructed a dataset composed of 1 410 246 tweets that we grouped in one big social graph. The data crawling was performed using NodeXL \cite{Smith} and took place during the end of 2020. 
Figure \ref{fig3} and Table \ref{tab3} showcase the main topics covered by the dataset.

\begin{table*}[ht]
	\centering
	\begin{tabular}{|p{0.15\textwidth}|p{0.75\textwidth}|}
		\hline
		\textbf{Main topic} & \textbf{Subtopics} \\
		\hline
		Computer science & Machine learning, Deep learning, Artificial intelligence, Big data, Graph database, Open Data, IoT, 5G, Social graph, Cyber security, Cyber-attack, Blockchain, Bitcoin, Hack, IBM, Data science, Power BI, Robotics, Smart city, Smart Home, Digital Predictive Analytics, Mathematics, Cisco, self-driving cars, VMware, Virtual reality, Web, domains, TensorFlow. \\
		\hline
		Politics & American express, Free speech, Black lives matter, Time is up, Immigration, Twitterstorian, Brexit UK, Vote, President Trump, Democracy, Breaking News, Democrats, Racism, white supremacy. \\
		\hline
		Health & Covid-19, flatern the curve, Vaccine, Cholera, intermittent fasting, Sugar free diet, Healthcare, HealthTech, Hemophilia, Malaria, Paludism, World Mustiquo day. Pregnancy, Abortion, protest against Exams in covid, diabetes, cannabis, AIDS, C-Section, Hydroxychloroquine, personalized Medicine. \\
		\hline
		Environment & Biodiversity, Food security, Climate change, Dogs, Dogs lovers, Fosil oil, fuels, Global warming, CO2, climate Strike. \\
		\hline
		Education & Books, E-books, QuickBooks, science teachers, Distance learning, Homeschooling, School closing, School reopening, Online learning, Book awards celebration. \\
		\hline
		Public figures & Bernie Senders, Joe Biden, Donald Trump, Michelle Obama, Snowden, Michael Cohen, Liam Payne, Bill Gates, SpaceX. \\
		\hline
		Social Media & Social Media, Social Media Marketing, Blogging, Tik Tok, Twitter, delete Facebook, YouTube, Podcast. \\
		\hline
		Marketing & Marketing, Social Media Marketing, Digital Marketing, public relations \\
		\hline
		Journalism & Journalism, sociologist, Forbs, Articles. \\
		\hline
		Various & Motor Trend, Unilever, Post Master, boycott whole food, Toxic masculinity, Feminism, Sexual harassments, save your children, lets chat, social pulse summit, youth day. \\
		\hline
	\end{tabular}
	\caption{Topics and subtopics covered the dataset}
	\label{tab3}
\end{table*}

\subsection{Adapted EHO for information foraging results}

\subsubsection{Empirical parameters setting}\label{EHOIFParam}
We conducted extensive tests for the sake of tuning the empirical parameters to values that ensure the best results in terms of relevance and response time, i.e. maximizing the similarity between the user's interests and the surfing path while minimizing the execution time. 
It is important to note that the stochastic aspect of the EHO algorithm requires to test each parameter value multiple times, to get stable outcomes. For that purpose, we run the tests at least 100 times for each parameter.

\begin{figure*}[!hbtp]
	\centering
	\begin{subfigure}{.4\textwidth}
		\centering
		\includegraphics[scale=0.4]{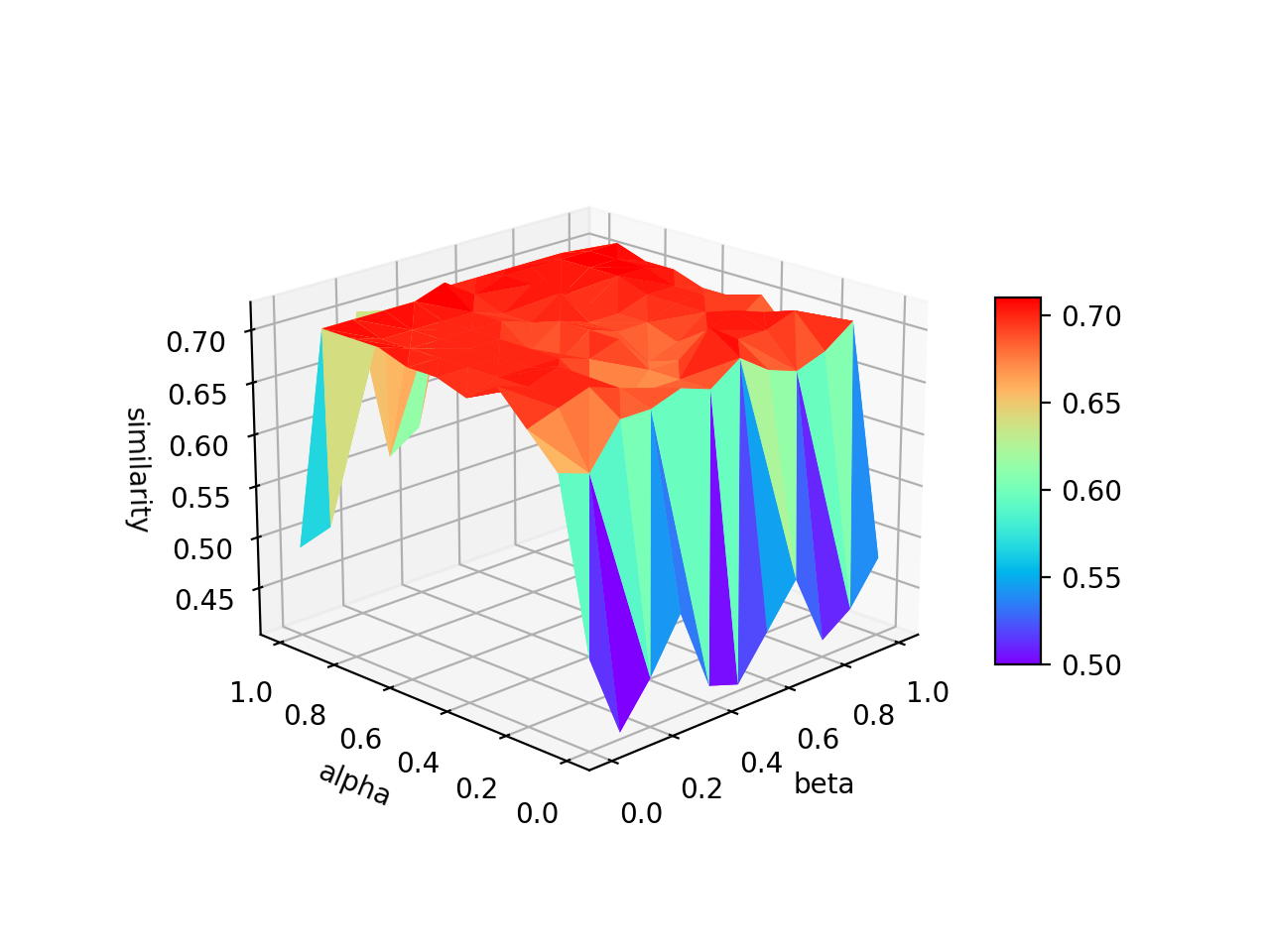}
		\caption{Similarity variation}
		\label{fig7a}
	\end{subfigure}%
	\begin{subfigure}{.4\textwidth}
		\centering
		\includegraphics[scale=0.4]{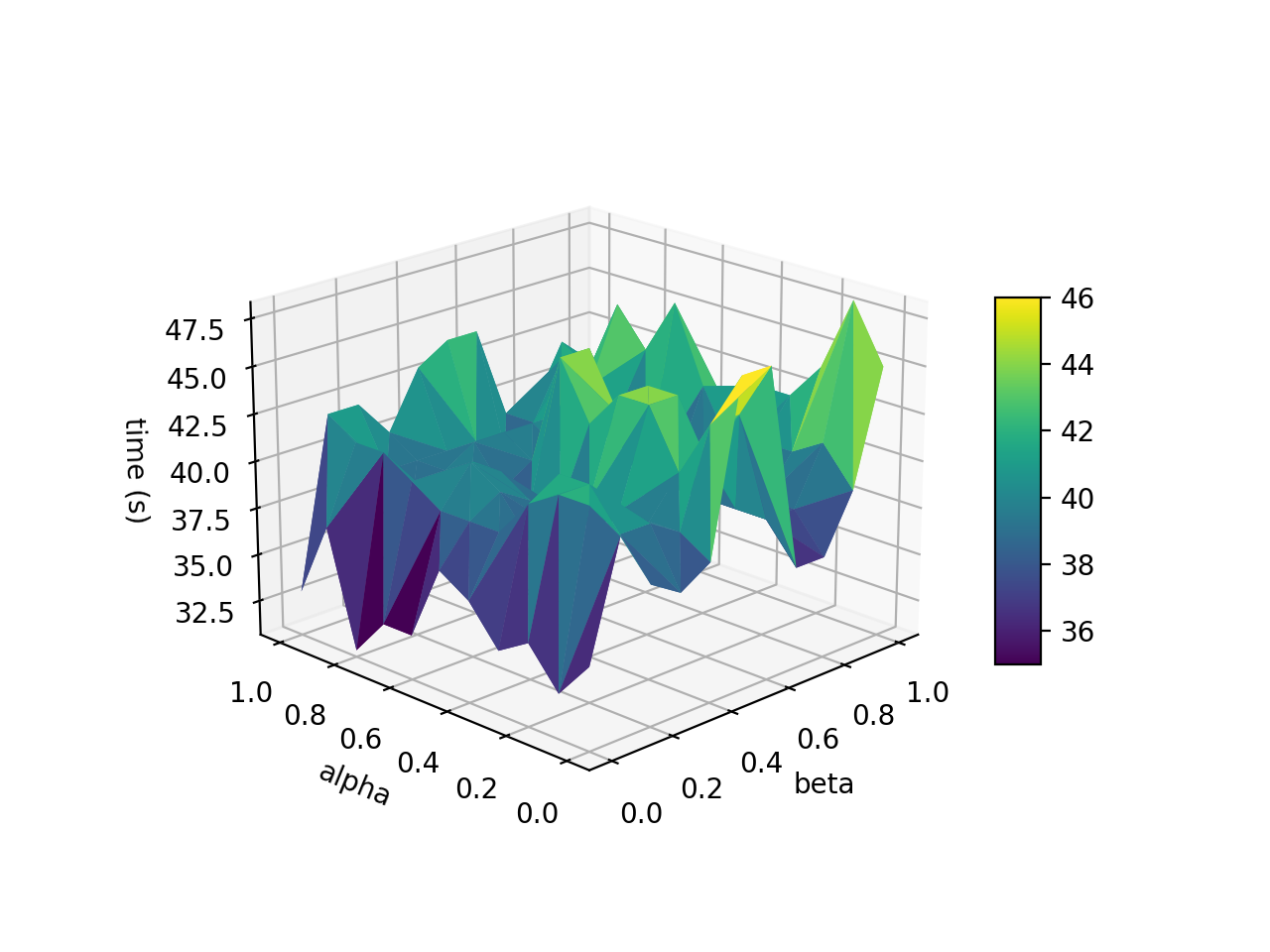}
		\caption{Time variation}
		\label{fig7b}
	\end{subfigure}
	\caption{Setting $\alpha$ and $\beta$ parameters based on Time and Similarity variation}
	\label{fig7}
\end{figure*}

First, we started with parameters $\alpha \in [0, 1]$ and $\beta \in [0, 1]$. Recall that $\alpha$ is a scale parameter that  determines the influence of the matriarch's position on rest of the elephants of the same clan, while $\beta$ determines the influence of the average position of the clan on the matriarch's position.
To select the best values of both parameters, we combined each value of $\alpha$ in the range $[0, 1]$ with all possible values of $\beta$ also in the same range.
Figure \ref{fig7a} shows the similarity score results, while  Figure\ref{fig7b} displays the response time results in seconds. The 3D representation gives a good visualization of the similarity and time evolution with the variation of the parameters. With respect to the results showed in both figures we set $\alpha$ to $0.9$ and $\beta$ to $0.4$.

\begin{figure*}[!hbtp]
	\centering
	\begin{subfigure}{.4\textwidth}
		\centering
		\includegraphics[scale=0.4]{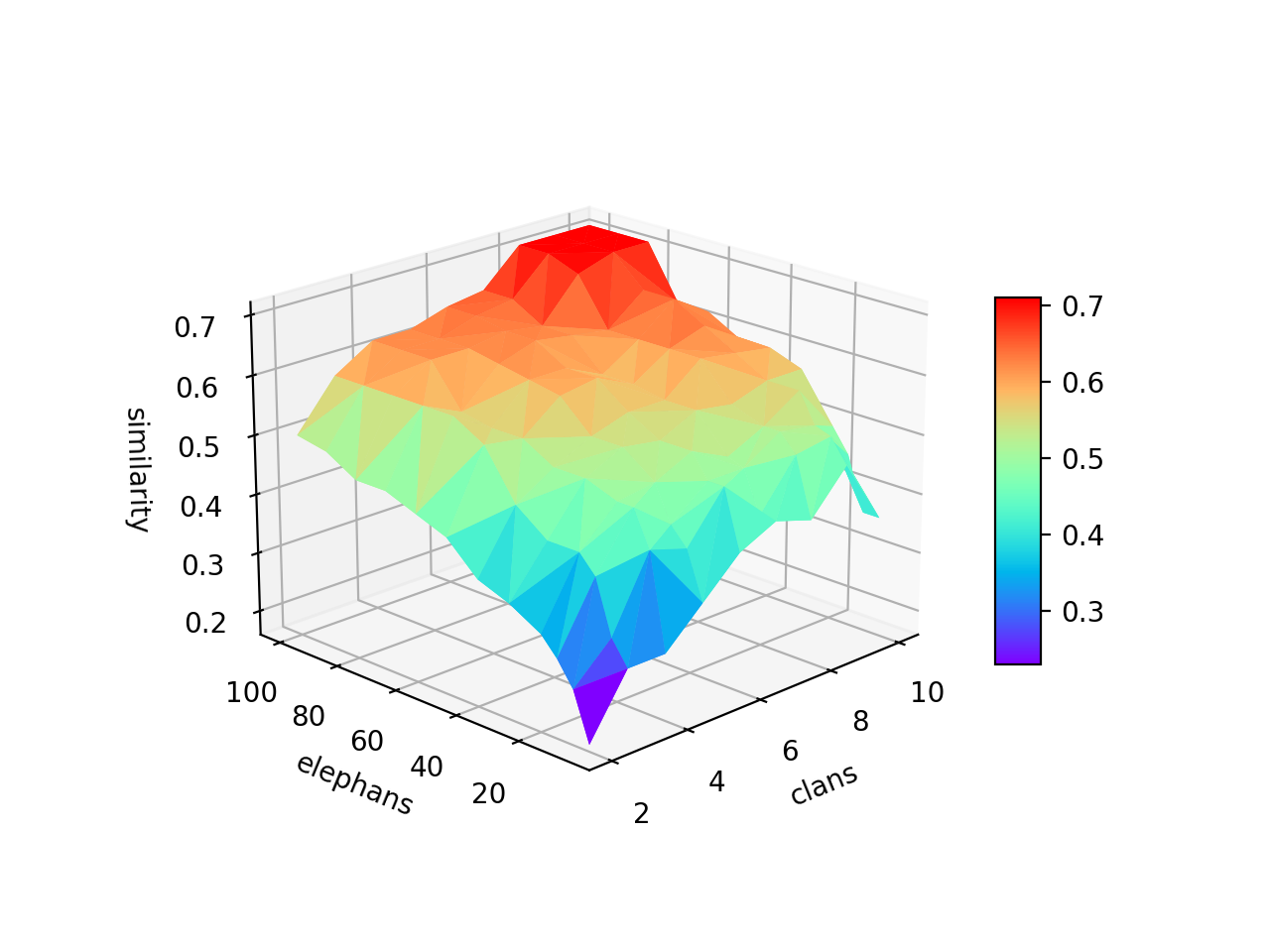}
		\caption{Similarity variation}
		\label{fig8a}
	\end{subfigure}%
	\begin{subfigure}{.4\textwidth}
		\centering
		\includegraphics[scale=0.4]{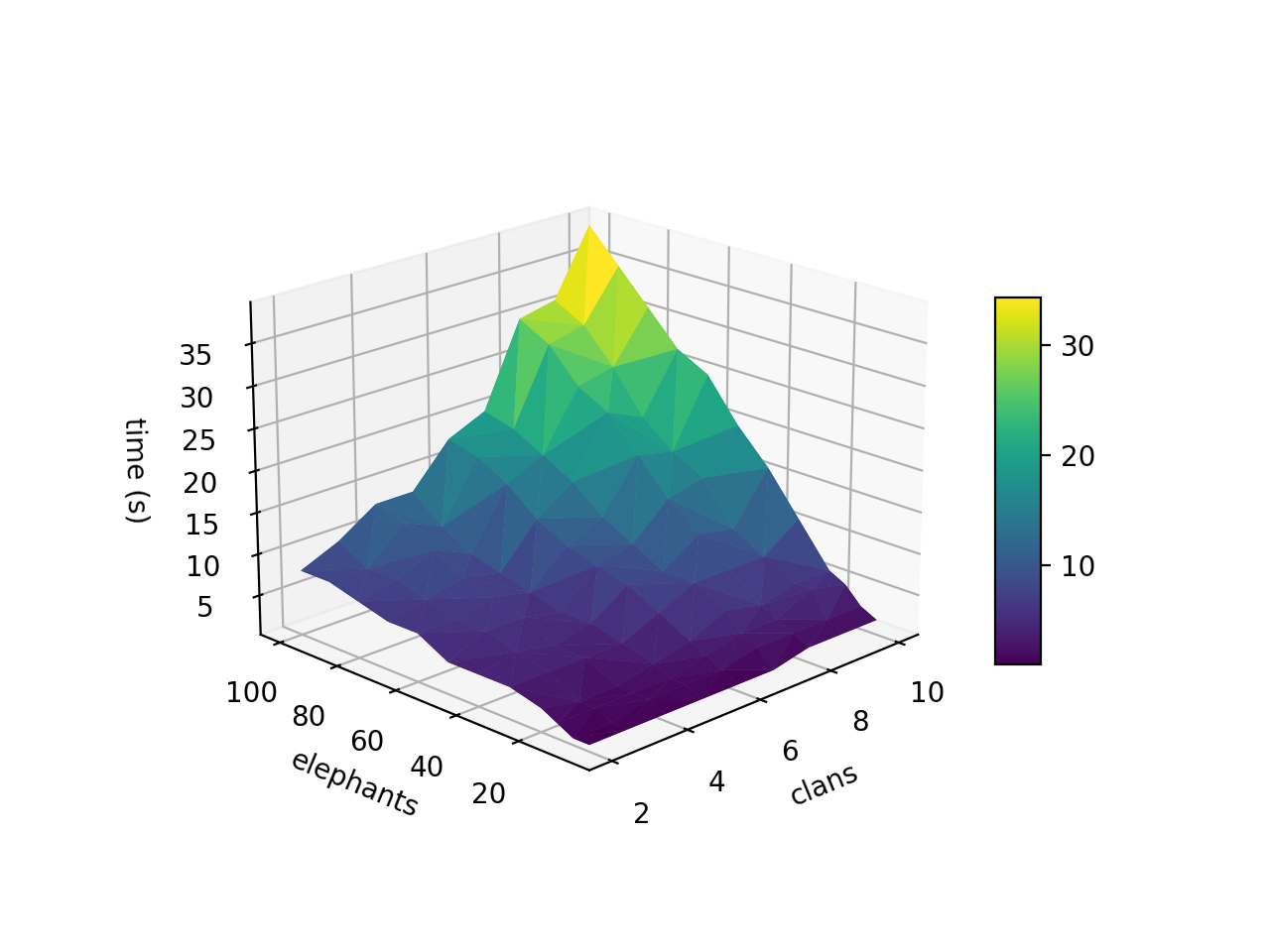}
		\caption{Time variation}
		\label{fig8b}
	\end{subfigure}
	\caption{Setting the number of clans and elephants based on Time and Similarity variation}
	\label{fig8}
\end{figure*}

Another important combination of parameters is the number of clans and the number of elephants in each clan. A proper number will help to visit different parts of the social graph and therefore get closer to relevant posts.
The results shown in Figure \ref{fig8a} and Figure \ref{fig8b} allow to determine the adequate number of clans and the number of elephants in each clan. The number of clans is set to $8$, with $90$ elephants in each clan.

\begin{figure*}[!hbtp]
	\centering
	\begin{subfigure}{.4\textwidth}
		\centering
		\includegraphics[scale=0.35]{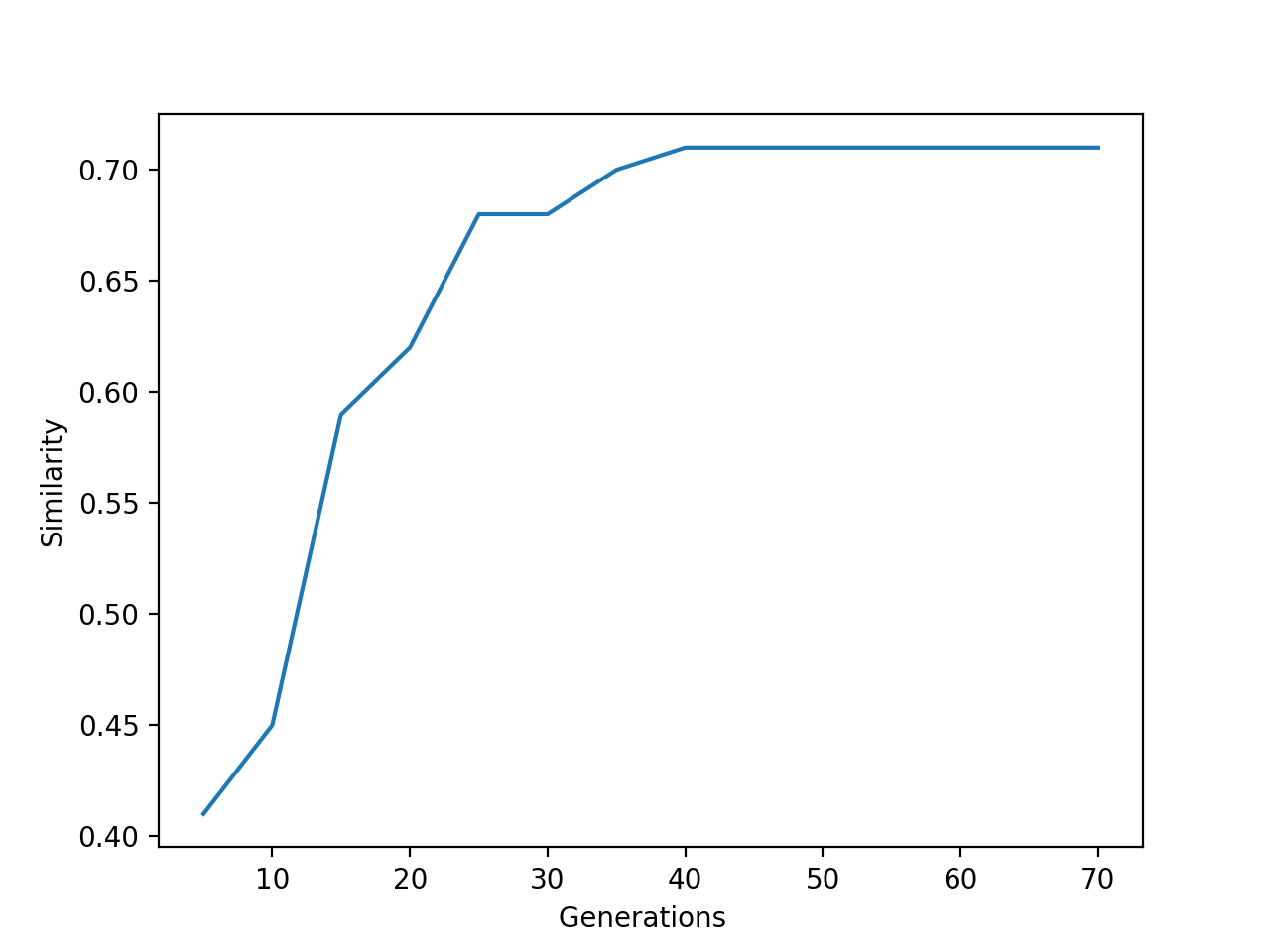}
		\caption{Similarity variation}
		\label{fig9a}
	\end{subfigure}%
	\begin{subfigure}{.4\textwidth}
		\centering
		\includegraphics[scale=0.35]{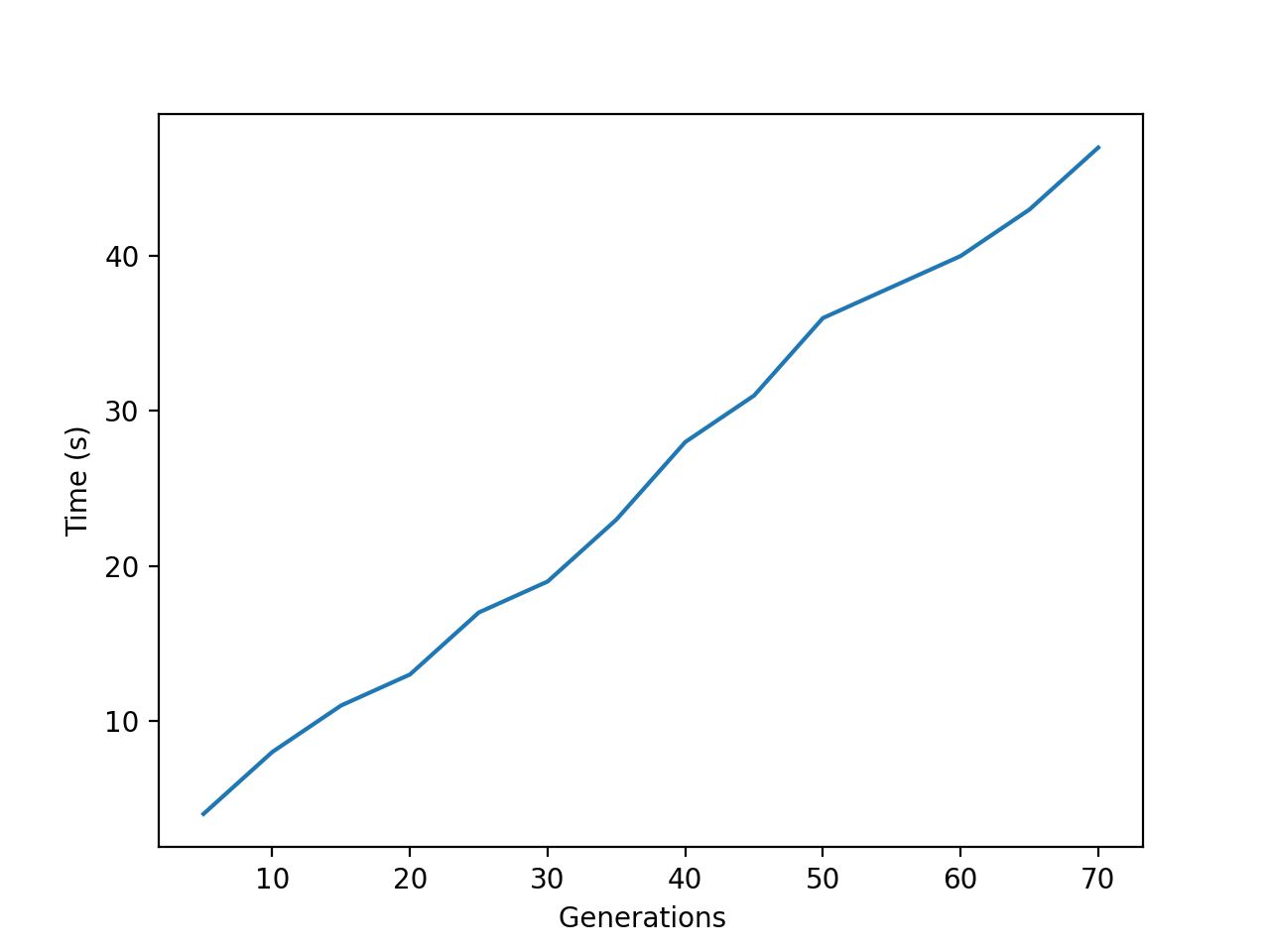}
		\caption{Time variation}
		\label{fig9b}
	\end{subfigure}
	\caption{Setting the number of generations based on Time and Similarity variation}
	\label{fig9}
\end{figure*}

The number of generations is the parameter that allows the algorithm to evolve a sufficient amount of time so it can reach better results and approach the global optimum.
We can observe from Figure \ref{fig9} that the best number of generations would be 40, since it maximizes the similarity and minimizes the response time.

\subsubsection{Foraging results }
Table \ref{tab2} presents some examples of the adapted EHO for information foraging results with 7 different users' interests (column one) generated for evaluation purpose. 
The surfing path with the most relevant tweet is displayed in column two, the similarity value between the surfing path and the user's interests is shown in column three, and the response time in seconds alongside the length of the surfing path are displayed in columns four and five, respectively. Note that when the surfing depth is greater than 1, the entire surfing path is displayed in chronological order of access, as in the case of the user's interest "diabetes type 2, intermittent fasting," for example.

\begin{table*}[!hbtp]
	\centering
	\begin{tabular}{|c|c|c|c|c|}
		\hline
		\textit{\textbf{User's interests}}                                                                & \textit{\textbf{Most relevant surfing path}}                                                                                                                                                                                                              & \textit{\textbf{Score}} & \textit{\textbf{Time (s)}} & \textit{\textbf{Surfing depth}} \\ \hline
		\begin{tabular}[c]{@{}c@{}}Machine Learning, \\ IA, Python\end{tabular}                           & \begin{tabular}[c]{@{}c@{}}Python for Machine Learning and Data Mining \\ \#DeepLearning \#datamining \#learning via \\ https://t.co/qcC4wrx6m6 \\  https://t.co/kLvs68HzEQ\end{tabular}                                                                  & 0.71                    & 26                         & 1                               \\ \hline
		\begin{tabular}[c]{@{}c@{}}American Express, \\ free speech, democracy\end{tabular}               & \begin{tabular}[c]{@{}c@{}}American Express \\ https://t.co/o9suYDdsV3\end{tabular}                                                                                                                                                                       & 0.64                    & 27                         & 1                               \\ \hline
		\begin{tabular}[c]{@{}c@{}}Public Relations, \\ communication\end{tabular}                        & \begin{tabular}[c]{@{}c@{}}A public relations strategy is critical \\ now more than ever \#PublicRelations \\ https://t.co/KZmGOD2Xjg\end{tabular}                                                                                                        & 0.54                    & 29                         & 1                               \\ \hline
		\begin{tabular}[c]{@{}c@{}}COVID19 immunity \\ transmission\end{tabular}                          & \begin{tabular}[c]{@{}c@{}}@CoocoLa\_Vrej WHO is still not sure if those \\ who recovered from COVID 19 develop a \\ certain immunity that they will not get \\ COVID 19 virus again.\end{tabular}                                                        & 0.57                    & 26                         & 1                               \\ \hline
		\multirow{3}{*}{\begin{tabular}[c]{@{}c@{}}diabetes type 2, \\ intermittent fasting\end{tabular}} & \begin{tabular}[c]{@{}c@{}}Can intermittent fasting make \\ you diabetic?\end{tabular}                                                                                                                                                                    & \multirow{3}{*}{0.74}   & \multirow{3}{*}{28}        & \multirow{3}{*}{3}              \\ \cline{2-2}
		& \begin{tabular}[c]{@{}c@{}}Does anyway here do intermittent fasting? \\ How do you do it?\end{tabular}                                                                                                                                                    &                         &                            &                                 \\ \cline{2-2}
		& \begin{tabular}[c]{@{}c@{}}Intermittent fasting has proven to help cure \\ Type II diabetes\end{tabular}                                                                                                                                                  &                         &                            &                                 \\ \hline
		\begin{tabular}[c]{@{}c@{}}Digital marketing, \\ business, social media\end{tabular}              & \begin{tabular}[c]{@{}c@{}}RT @V2M2Group: Get Social: The Power of \\ Social Media for Marketing Your Business?\\ \#business \#digitalmarketing \#marketing \\ \#smallbusiness \#SocialMedia \\ \#GuernseyBusinesses https://t.co/8wnlALHXsh\end{tabular} & 0.73                    & 29                         & 1                               \\ \hline
		Bitcoin prices market                                                                             & \begin{tabular}[c]{@{}c@{}}Bitcoin price within about 3\% of gold price \\ https://t.co/GwjcMSB9Jp\end{tabular}                                                                                                                                           & 0.67                    & 24                         & 1                               \\ \hline
		\multirow{3}{*}{\begin{tabular}[c]{@{}c@{}}Joe Biden and \\ Bernie Senders\end{tabular}}          & \begin{tabular}[c]{@{}c@{}}@LyndaMo85130479 @BugOffDear Biden \\ positions are literally just\\ copy/pasted from Bernie Sanders\end{tabular}                                                                                                              & \multirow{3}{*}{0.46}   & \multirow{3}{*}{26}        & \multirow{3}{*}{3}              \\ \cline{2-2}
		& \begin{tabular}[c]{@{}c@{}}Folks mention Biden's past plagiarism\\ True\end{tabular}                                                                                                                                                                      &                         &                            &                                 \\ \cline{2-2}
		& \begin{tabular}[c]{@{}c@{}}But who believes Joe had anything to do\\ with deciding this, or preparing the doc?\\ Who is in charge?\\ https://t.co/x8RBCz4H6D\end{tabular}                                                                                 &                         &                            &                                 \\ \hline
		Smart City, 5G, IoT                                                                               & \begin{tabular}[c]{@{}c@{}}Samsung IoT Smart City\\ https://t.co/Xnf5JHnOq9\\  via @YouTube\\ @\_funtastic5\_ \#TelkomFuntastic5 \\ \#RWSTREG5 \#smartcity\end{tabular}                                                                                   & 0.70                    & 25                         & 1                               \\ \hline
	\end{tabular}
	\caption{Information Foraging Results}
	\label{tab2}
\end{table*}

We observe that in almost all cases, the system is capable of finding relevant tweets. However, the response time is relatively long, mainly because of the big size of the social graph and the fact that the foraging process happens exclusively online. 
We can also notice that the surfing depth is to a certain extent small, which can be explained by the fact that the social graph is not strongly connected. Moreover, during the construction of the surfing path, a tweet is only inserted if it is more relevant than the tweets that were accessed before it in the same path.

\subsection{Enhanced EHO for large scale information foraging results}
Although we were able to reach relevant posts using our first attempt based on the adaptation of the original EHO algorithm to information foraging, the results showed some limitations related to the efficiency, especially when it comes to big social graphs. To cope with this issue, we proposed in Section \ref{sec6} a novel approach consisting in an enhanced version of EHO for large scale infomration foraging. 

\subsubsection{Empirical parameters setting}
The first parameter to define is the number of territories, i.e. the number of clusters $k$. For this purpose, we tested the k-means algorithm with different values of $k$ in the interval $[1, 80]$. For each fixed number of clusters $k$, we use Formula (\ref{eq:wss}) to compute the total Within Cluster Sums of Squares (WSS), which measures the average distance between the posts and their corresponding centroids for each cluster \cite{Fauzan, Estiri}.

\begin{equation}
	WSS = \sum\limits_{i=1}^k \sum\limits_{\tilde{e} \in S_i} d(\tilde{e}, m_i)
	\label{eq:wss}
\end{equation}
Where:
\begin{itemize}
	\item $k$: is the number of clusters
	\item $S_i$: is a cluster
	\item $\tilde{e}$: is a post
	\item $m_i$ is the centroid of cluster $S_i$
	\item $d(\tilde{e}, m_i)$: is the euclidean distance between the post and its associated centroid
\end{itemize}

\begin{figure}[!htbp]
	\centering
	\includegraphics[scale=0.3]{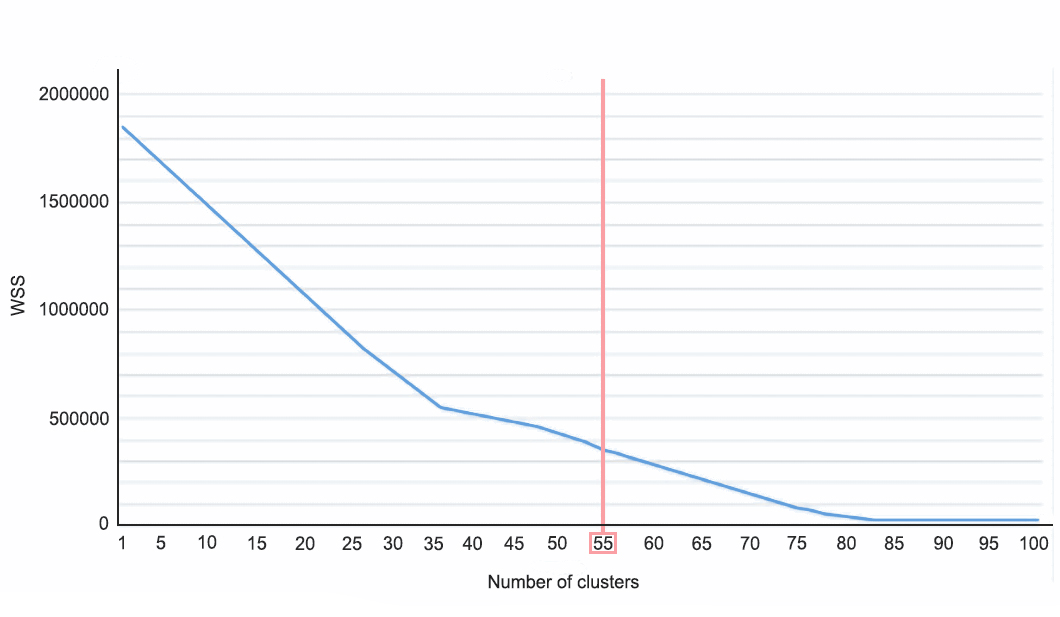}
	\caption{Within Cluster Sums of Squares plot}
	\label{fig10}
\end{figure}

Once the calculations are finished, we plot the curve of WSS according to the number of clusters $k$. The location of a bend (knee) in the plot is generally considered as an indicator for the proper number of clusters. The results shown in Figure \ref{fig10}, indicate that the best number of clusters is $k=55$. 

\begin{figure}[!htbp]
	\centering
	\includegraphics[scale=0.5]{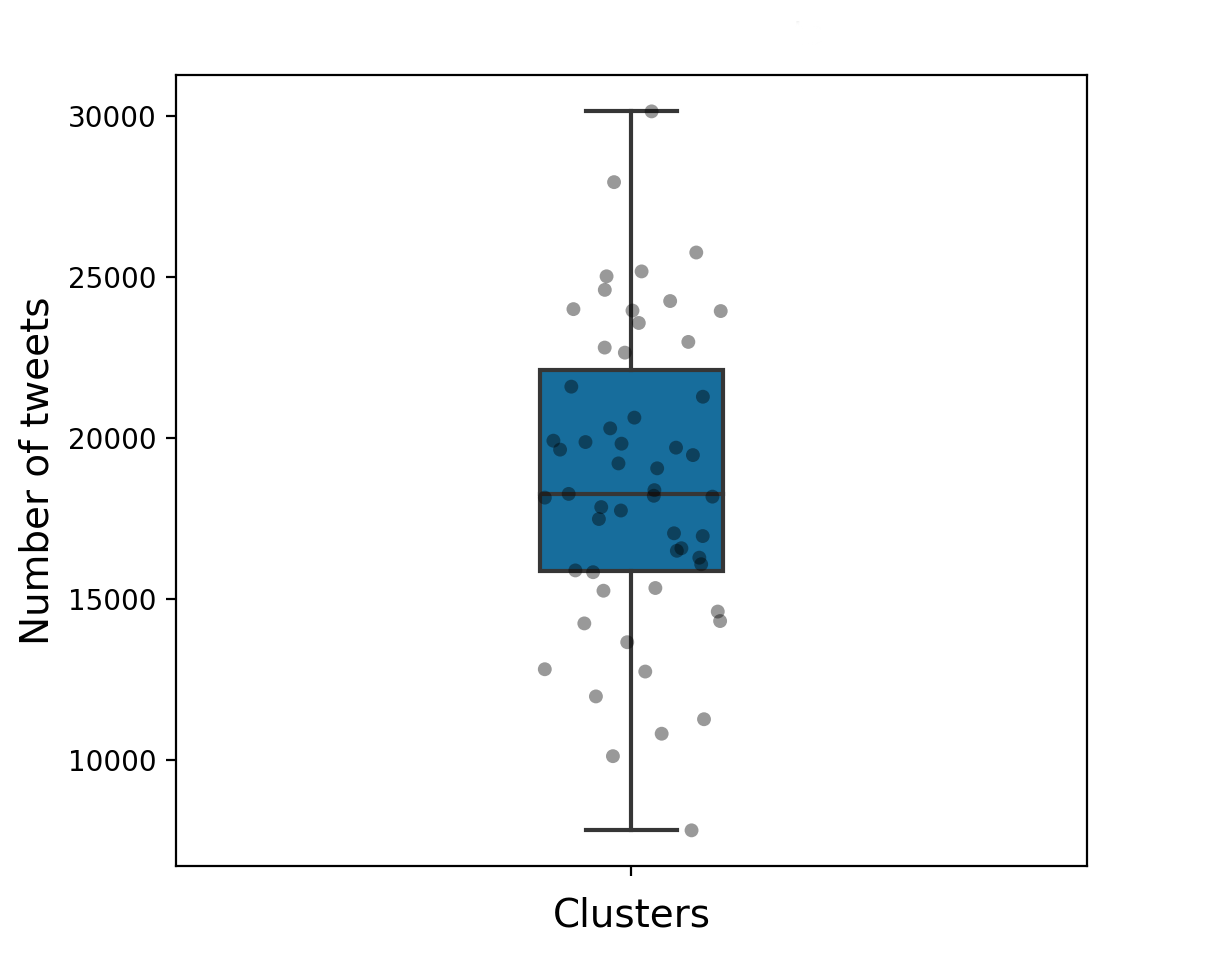}
	\caption{Boxplot of clustering results}
	\label{fig11}
\end{figure}

Figure \ref{fig11} displays the distribution of the 1 410 246 tweets over the 55 clusters. We observe that the smallest cluster contains 7 821 tweets while the largest one groups a total of 30 144 tweets with a median of 18 267 tweets per cluster. 

\begin{figure*}[!htbp]
	\centering
	\begin{subfigure}{.45\textwidth}
		\centering
		\includegraphics[scale=0.4]{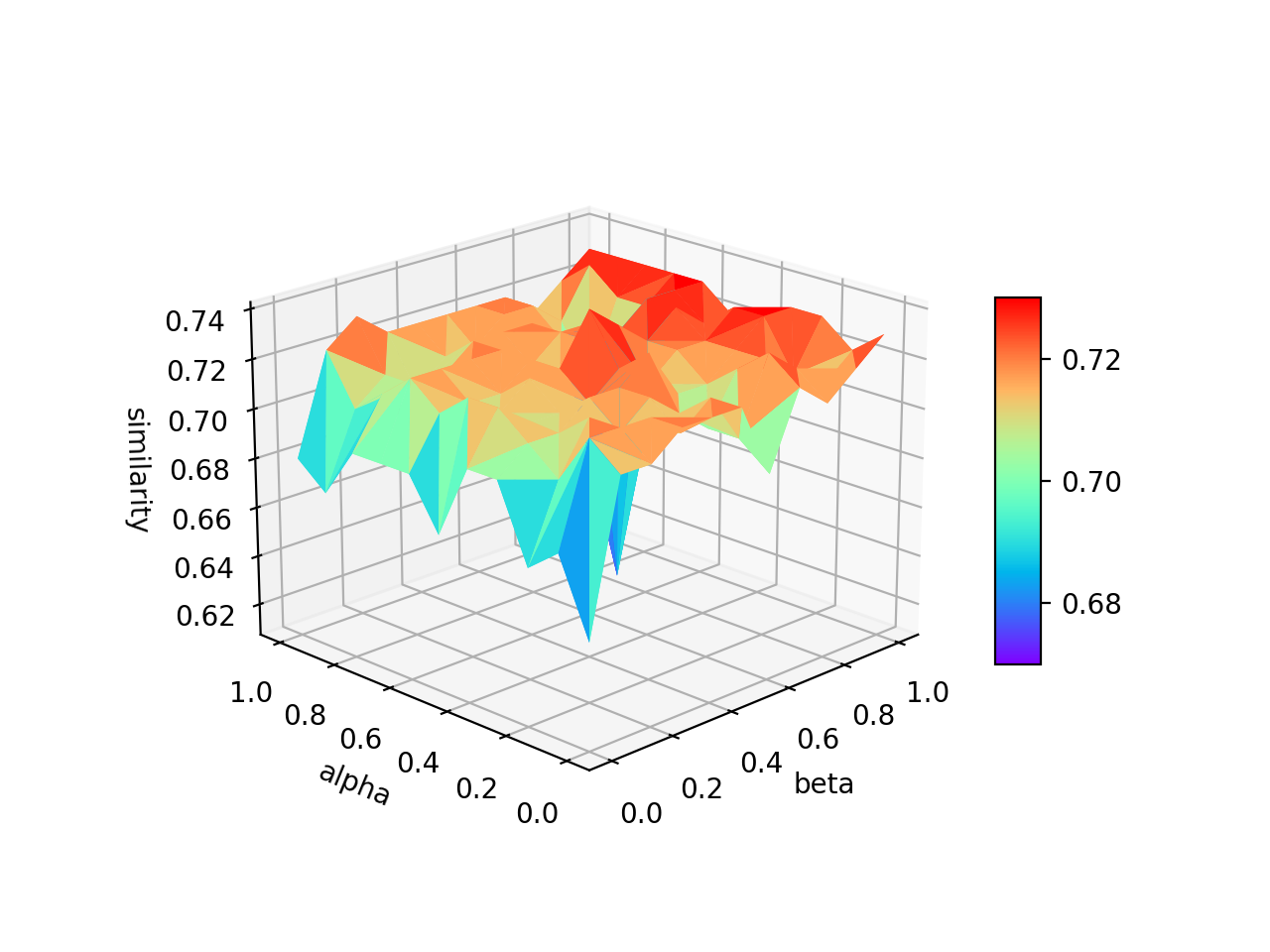}
		\caption{Similarity variation regarding $\alpha$ and $\beta$}
		\label{fig12a}
	\end{subfigure}%
	\begin{subfigure}{.45\textwidth}
		\centering
		\includegraphics[scale=0.4]{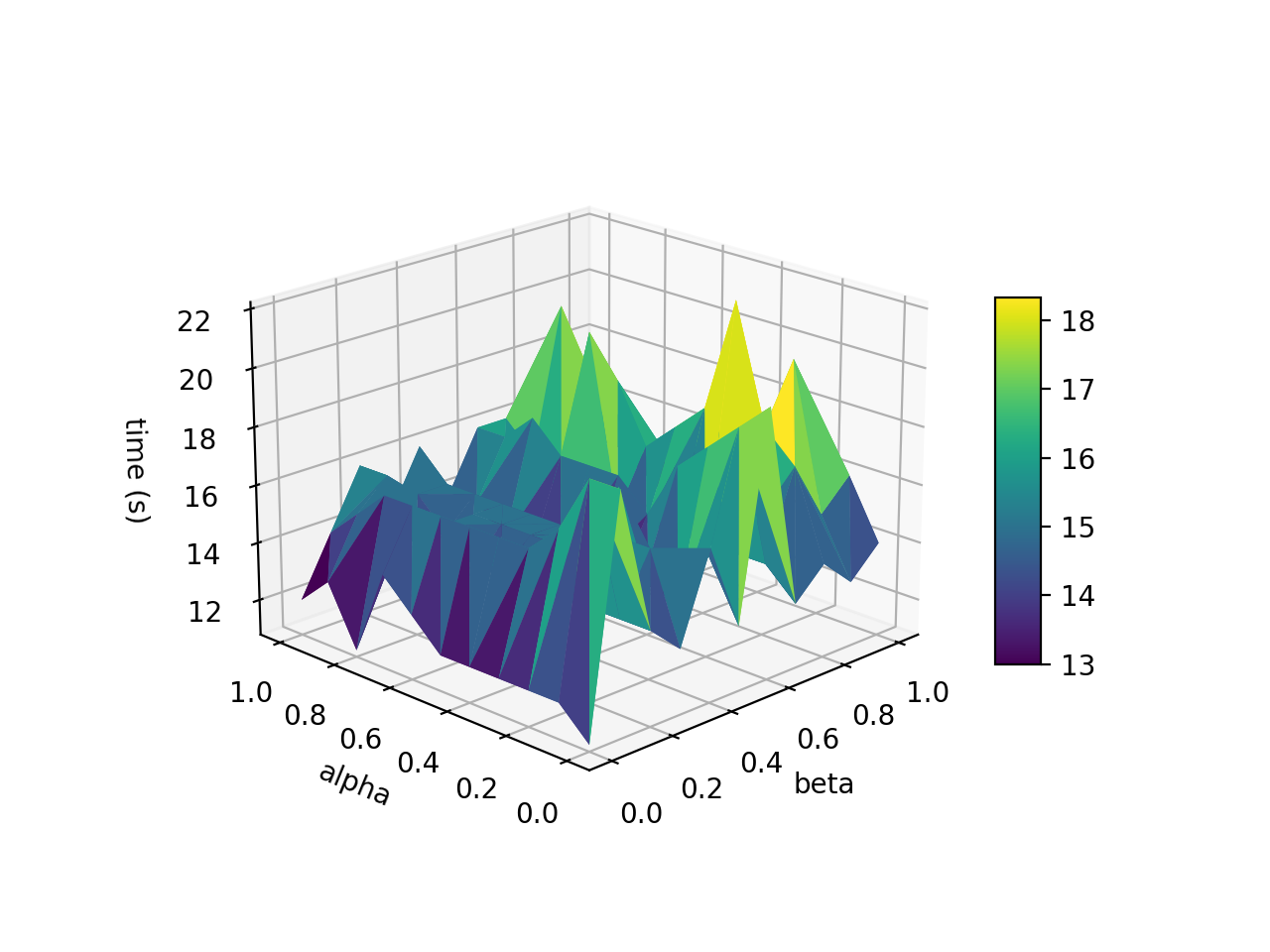}
		\caption{Time variation regarding $\alpha$ and $\beta$}
		\label{fig12b}
	\end{subfigure}
	\\
	\begin{subfigure}{.45\textwidth}
		\centering
		\includegraphics[scale=0.4]{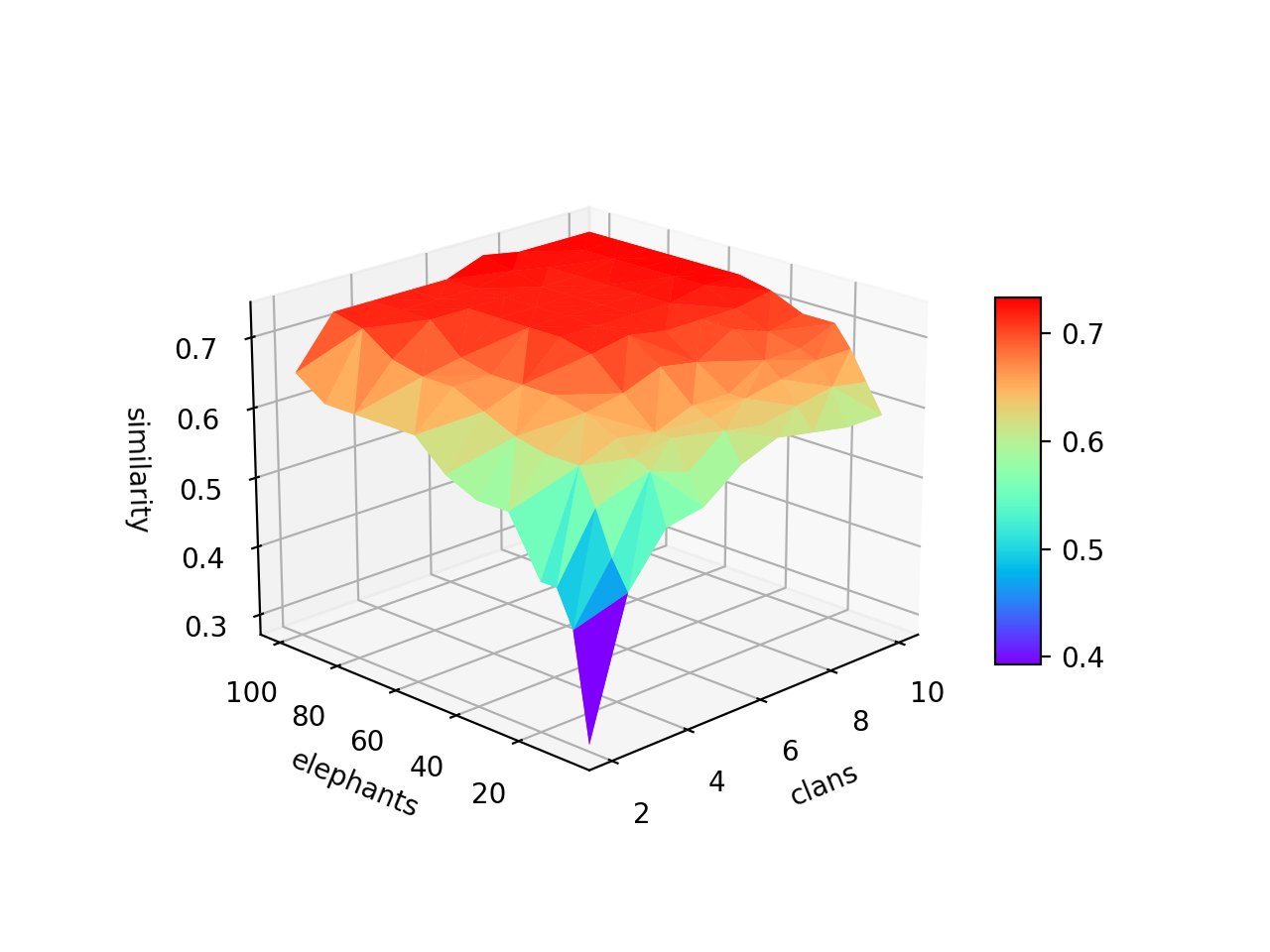}
		\caption{Similarity variation regarding the number of clans and elephants}
		\label{fig12c}
	\end{subfigure}%
	\begin{subfigure}{.45\textwidth}
		\centering
		\includegraphics[scale=0.4]{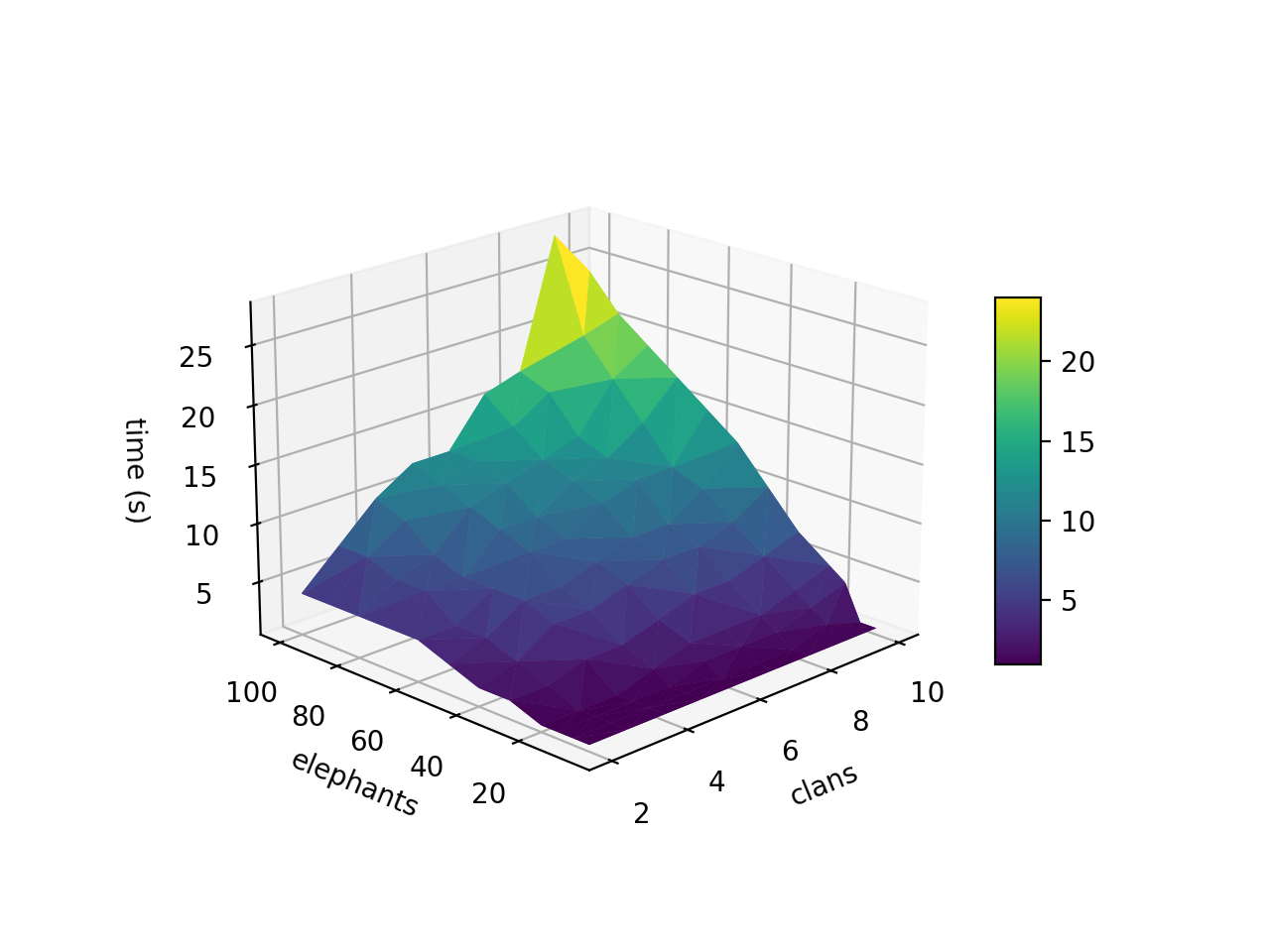}
		\caption{Time variation regarding the number of clans and elephants}
		\label{fig12d}
	\end{subfigure}%
	\\
	\begin{subfigure}{.45\textwidth}
		\centering
		\includegraphics[scale=0.35]{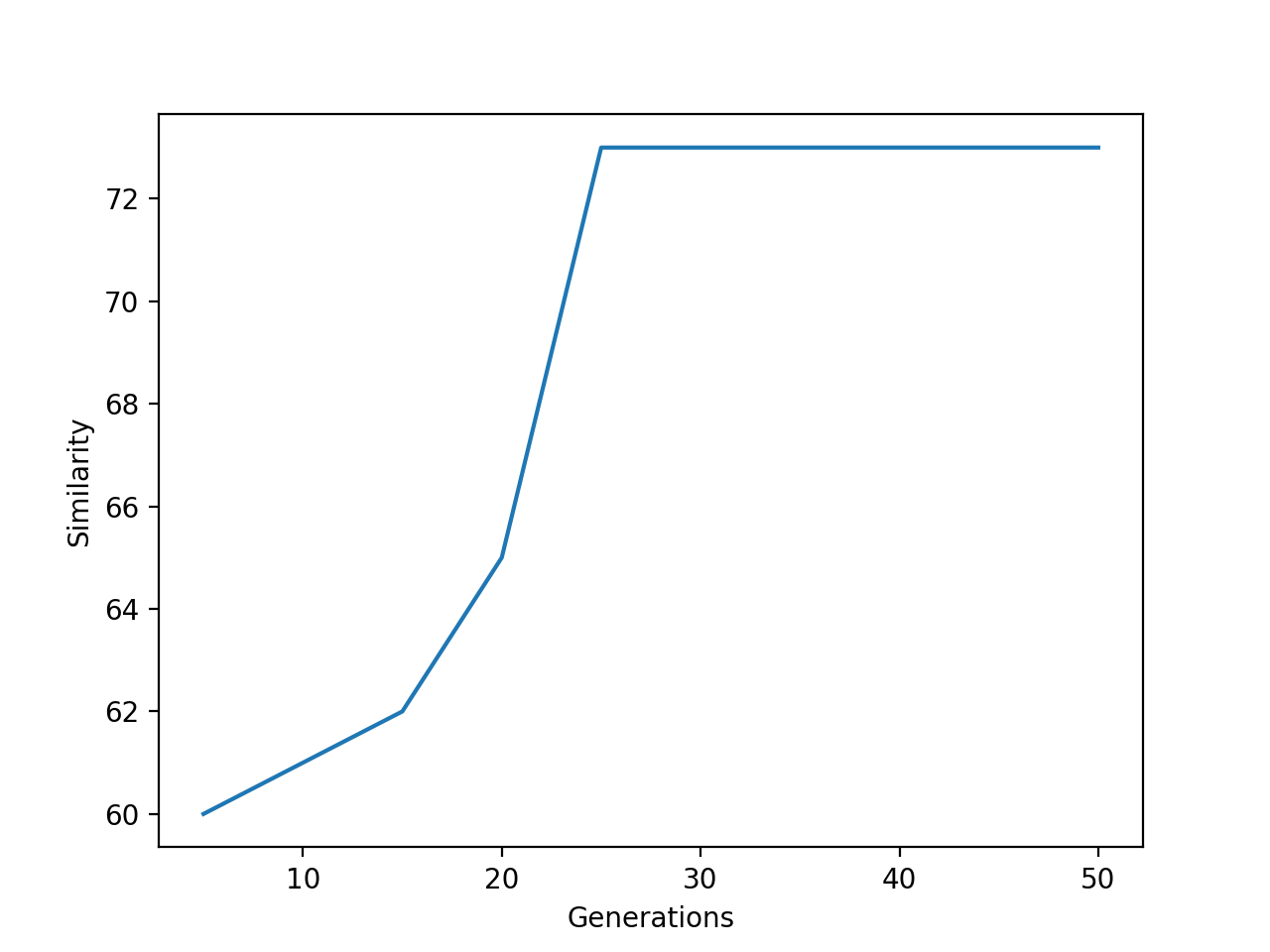}
		\caption{Similarity variation regarding the number of generations}
		\label{fig12e}
	\end{subfigure}%
	\begin{subfigure}{.45\textwidth}
		\centering
		\includegraphics[scale=0.35]{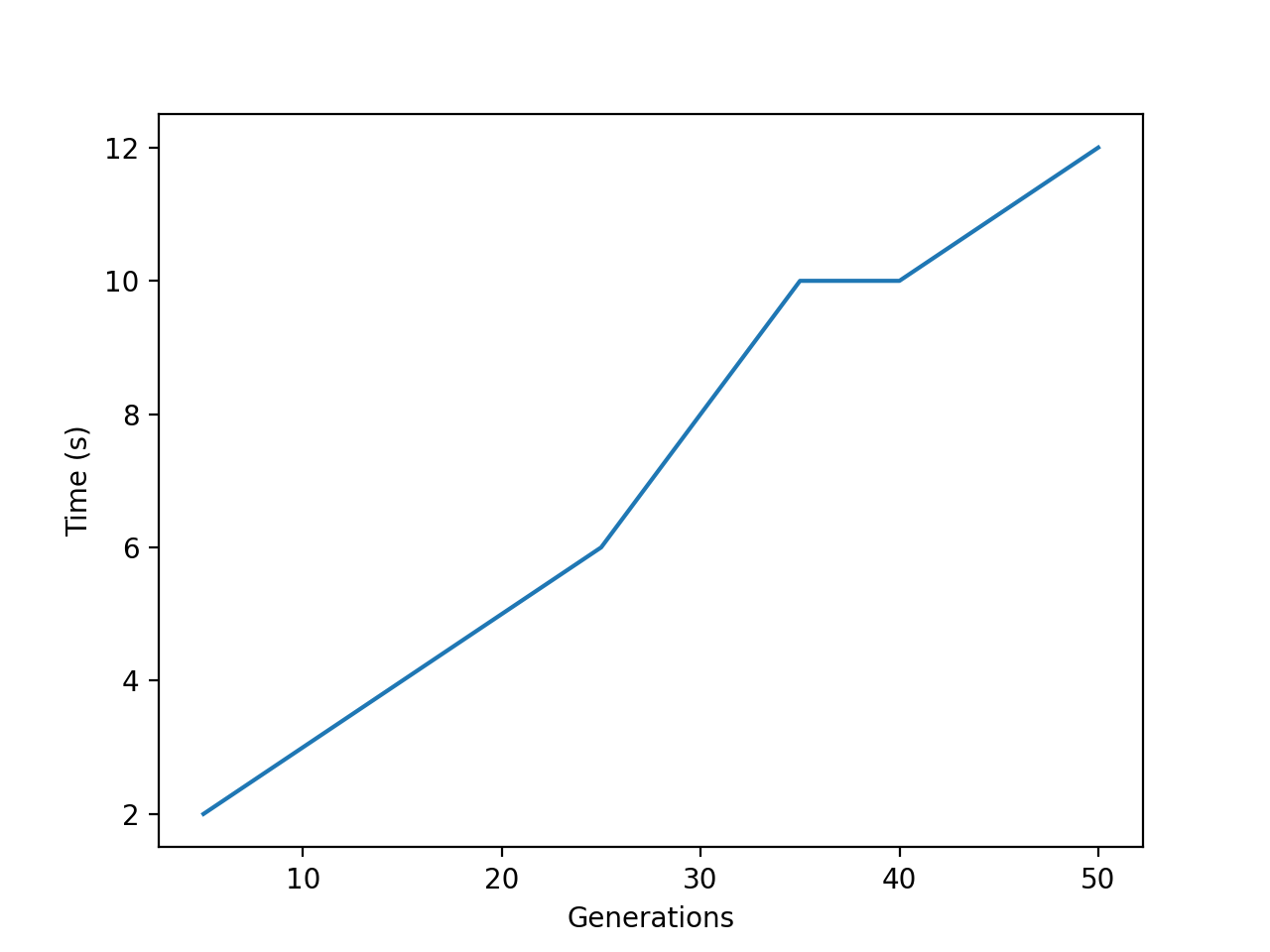}
		\caption{Time variation regarding the number of generations}
		\label{fig12f}
	\end{subfigure}%
	\caption{Setting the empirical parameters for enhanced EHO for large scale information foraging}
	\label{fig12}
\end{figure*}

Given that the social graph is now divided into 55 territories, the rest of the parameters needs to be tuned again accordingly. To do so, we conducted extensive tests following the same steps of subsection \ref{EHOIFParam}. We also performed the tests at least 100 times for each parameter. Figure \ref{fig12} shows the tests we undertook to set the empirical parameters $\alpha$, $\beta$, number of clans, number of elephants and the maximum number of generations, while Table \ref{tab5} shows the optimal values of these parameters.

\begin{table}[!htbp]
	\centering
	\begin{tabular}{|c|c|}
		\hline 
		\textbf{Parameter} & \textbf{Value} \\ 
		\hline 
		$\alpha$ & $0.5$ \\ 
		\hline 
		$\beta$ & $0.5$ \\ 
		\hline 
		$nClans$ & $5$  \\ 
		\hline 
		$n_{c_i}$ & $50$ \\ 
		\hline 
		$MaxGen$ & $25$ \\ 
		\hline 
		$q_0$ & $0.75$ \\ 
		\hline 
		$t_0$ & $6$ \\ 
		\hline 
	\end{tabular} 
	\caption{Empirical parameters values for enhanced EHO for large scale information foraging}
	\label{tab5}
\end{table}

\subsubsection{Foraging results}
The comparison results between adapted EHO for information foraging (EHOIF) and enhanced EHO for large scale information foraging (EEHOLSIF) are reported in Table \ref{tab6}. The first column represents the user's interest, while the rest of the columns provide for each approach the most relevant surfing path, its similarity with the user's interests, the response time and the surfing depth. 

We observe that both approaches are capable of finding relevant tweets that can potentially satisfy the user's interests. However, the main difference between the two approaches resides in the score and the response time. In fact, we can see that EEHOLSIF can achieve a higher score in almost all cases. Furthermore, its response time is considerably faster. 
We believe that this gain in performance is the result of the improvements we brought to the algorithm to make it able to undertake large scale information foraging and in particular the territory concept and the migration mechanism. 

\begin{table*}[!htbp]
	\tiny
	\centering
	\begin{tabular}{ccccccccc}
		\cline{2-9}
		\multicolumn{1}{c|}{}                                                                                                            & \multicolumn{4}{c|}{\textbf{EHOIF}}                                                                                                                                                                                                            & \multicolumn{4}{c|}{\textbf{EHO-KM-IF}}                                                                                     \\ \hline
		\multicolumn{1}{|c|}{\textbf{User’s interests}}                                                                               & \multicolumn{1}{c|}{\textbf{Most relevant surfing path}}                                                                                                                                                                                                                                  & \multicolumn{1}{c|}{\textbf{Score}}          & \multicolumn{1}{c|}{\textbf{\begin{tabular}[c]{@{}c@{}}Time\\ (s)\end{tabular}}} & \multicolumn{1}{c|}{\textbf{\begin{tabular}[c]{@{}c@{}}Surfing\\ depth\end{tabular}}} & \multicolumn{1}{c|}{\textbf{Most relevant surfing path}}                                                                                                                                                                                                          & \multicolumn{1}{c|}{\textbf{Score}}          & \multicolumn{1}{c|}{\textbf{\begin{tabular}[c]{@{}c@{}}Time\\ (s)\end{tabular}}} & \multicolumn{1}{c|}{\textbf{\begin{tabular}[c]{@{}c@{}}Surfing\\ depth\end{tabular}}} \\ \hline
		\multicolumn{1}{|c|}{\begin{tabular}[c]{@{}c@{}}Machine\\ Learning, \\ IA, Python\end{tabular}}                               & \multicolumn{1}{c|}{\begin{tabular}[c]{@{}c@{}}Python for Machine Learning and\\ Data Mining \#DeepLearning \\ \#datamining \#learning via \\ https://t.co/qcC4wrx6m6\\ https://t.co/kLvs68HzEQ\end{tabular}}                                                                             & \multicolumn{1}{c|}{0.71}                  & \multicolumn{1}{c|}{26 s}                                                        & \multicolumn{1}{c|}{1}                                                                & \multicolumn{1}{c|}{\begin{tabular}[c]{@{}c@{}}Introduction To Machine \\ Learning with Python \\ \#MachineLearning \#deeplearning \\ \#learning via \\ https://t.co/lWfQGVjKXK\\ https://t.co/ZobXDvNWRO\end{tabular}}                                           & \multicolumn{1}{c|}{0.71}                  & \multicolumn{1}{c|}{1 s}                                                         & \multicolumn{1}{c|}{1}                                                                \\ \hline
		\multicolumn{1}{|c|}{\begin{tabular}[c]{@{}c@{}}American \\ Express, \\ free speech, \\ democracy\end{tabular}}               & \multicolumn{1}{c|}{\begin{tabular}[c]{@{}c@{}}American Express\\ https://t.co/o9suYDdsV3\end{tabular}}                                                                                                                                                                                   & \multicolumn{1}{c|}{0.64}                  & \multicolumn{1}{c|}{27s}                                                         & \multicolumn{1}{c|}{1}                                                                & \multicolumn{1}{c|}{\begin{tabular}[c]{@{}c@{}}American Express\\ https://t.co/bGaqvvZp69\end{tabular}}                                                                                                                                                           & \multicolumn{1}{c|}{0.64}                  & \multicolumn{1}{c|}{1.1s}                                                          & \multicolumn{1}{c|}{1}                                                                \\ \hline
		\multicolumn{1}{|c|}{\begin{tabular}[c]{@{}c@{}}Public \\ Relations, \\ communication\end{tabular}}                           & \multicolumn{1}{c|}{\begin{tabular}[c]{@{}c@{}}A public relations strategy is\\ critical now more than ever\\ \#PublicRelations \\ https://t.co/KZmGOD2Xjg\end{tabular}}                                                                                                                  & \multicolumn{1}{c|}{0.54}                  & \multicolumn{1}{c|}{29 s}                                                        & \multicolumn{1}{c|}{1}                                                                & \multicolumn{1}{c|}{\begin{tabular}[c]{@{}c@{}}RT: The Relevance \\ Of Public Relations \& \\ Communication In Fashion\\ https://t.co/YrOhuv0mwX\\ \#fashion \#Fashionista \\ \#bloggerstr\end{tabular}}                                                             & \multicolumn{1}{c|}{0.76}                  & \multicolumn{1}{c|}{0.8s}                                                          & \multicolumn{1}{c|}{1}                                                                \\ \hline
		\multicolumn{1}{|c|}{\begin{tabular}[c]{@{}c@{}}COVID19 \\ immunity \\ transmission\end{tabular}}                             & \multicolumn{1}{c|}{\begin{tabular}[c]{@{}c@{}}@CoocoLa\_Vrej WHO is \\ still not sure if those \\ who recovered from \\ COVID 19 develop a \\ certain immunity that \\ they will not get \\ COVID 19 virus again.\end{tabular}}                                                          & \multicolumn{1}{c|}{0.57}                  & \multicolumn{1}{c|}{26 s}                                                        & \multicolumn{1}{c|}{1}                                                                & \multicolumn{1}{c|}{\begin{tabular}[c]{@{}c@{}}@CoocoLa\_Vrej WHO is \\ still not sure if those\\ who recovered from\\ COVID 19 develop a\\ certain immunity that\\ they will not get\\ COVID 19 virus again.\end{tabular}}                                       & \multicolumn{1}{c|}{0.57}                  & \multicolumn{1}{c|}{1.2s}                                                          & \multicolumn{1}{c|}{1}                                                                \\ \hline
		\multicolumn{1}{|c|}{\multirow{3}{*}{\begin{tabular}[c]{@{}c@{}}diabetes \\ type 2, \\ intermittent \\ fasting\end{tabular}}} & \multicolumn{1}{c|}{\begin{tabular}[c]{@{}c@{}}Can intermittent fasting \\ make you diabetic?\end{tabular}}                                                                                                                                                                               & \multicolumn{1}{c|}{\multirow{3}{*}{0.74}} & \multicolumn{1}{c|}{\multirow{3}{*}{28s}}                                        & \multicolumn{1}{c|}{\multirow{3}{*}{3}}                                               & \multicolumn{1}{c|}{\multirow{3}{*}{\begin{tabular}[c]{@{}c@{}}RT @EvolveHolistic: How \\ to Intermittent Fast and\\ Which Type of Fasting\\ Is Right for You\\ https://t.co/sx9iNQr312\\  https://t.co/v2QVQVVRcv\end{tabular}}}                                 & \multicolumn{1}{c|}{\multirow{3}{*}{0.73}} & \multicolumn{1}{c|}{\multirow{3}{*}{1s}}                                         & \multicolumn{1}{c|}{\multirow{3}{*}{1}}                                               \\ \cline{2-2}
		\multicolumn{1}{|c|}{}                                                                                                        & \multicolumn{1}{c|}{\begin{tabular}[c]{@{}c@{}}Does anyway here do\\ intermittent fasting? \\ How do you do it?\end{tabular}}                                                                                                                                                             & \multicolumn{1}{c|}{}                      & \multicolumn{1}{c|}{}                                                            & \multicolumn{1}{c|}{}                                                                 & \multicolumn{1}{c|}{}                                                                                                                                                                                                                                             & \multicolumn{1}{c|}{}                      & \multicolumn{1}{c|}{}                                                            & \multicolumn{1}{c|}{}                                                                 \\ \cline{2-2}
		\multicolumn{1}{|c|}{}                                                                                                        & \multicolumn{1}{c|}{\begin{tabular}[c]{@{}c@{}}Intermittent fasting has proven \\ to help cure Type II diabetes\end{tabular}}                                                                                                                                                             & \multicolumn{1}{c|}{}                      & \multicolumn{1}{c|}{}                                                            & \multicolumn{1}{c|}{}                                                                 & \multicolumn{1}{c|}{}                                                                                                                                                                                                                                             & \multicolumn{1}{c|}{}                      & \multicolumn{1}{c|}{}                                                            & \multicolumn{1}{c|}{}                                                                 \\ \hline
		\multicolumn{1}{|c|}{\begin{tabular}[c]{@{}c@{}}Digital \\ marketing,\\ business, \\ social media\end{tabular}}               & \multicolumn{1}{c|}{\begin{tabular}[c]{@{}c@{}}RT @V2M2Group: Get \\ Social: The Power of \\ Social Media for Marketing\\  Your Business ?\\ \#business \#digitalmarketing \\ \#marketing \#smallbusiness \\ \#SocialMedia\\ \#GuernseyBusinesses\\ https://t.co/8wnlALHXsh\end{tabular}} & \multicolumn{1}{c|}{0.73}                  & \multicolumn{1}{c|}{29s}                                                         & \multicolumn{1}{c|}{1}                                                                & \multicolumn{1}{c|}{\begin{tabular}[c]{@{}c@{}}Learn How to Market \\ Your Business on Social\\ Media -- Affiliate or \\ Network Marketing on\\ Social Media \\ https://t.co/iK9SrVzuLM\\  \#OnlineBusiness \#SocialMedia\\ https://t.co/45waDGLoGh\end{tabular}} & \multicolumn{1}{c|}{0.76}                  & \multicolumn{1}{c|}{1s}                                                          & \multicolumn{1}{c|}{1}                                                                \\ \hline
		\multicolumn{1}{|c|}{\begin{tabular}[c]{@{}c@{}}Bitcoin \\ prices \\ market\end{tabular}}                                     & \multicolumn{1}{c|}{\begin{tabular}[c]{@{}c@{}}Bitcoin price within about \\ 3\% of gold price\\ https://t.co/GwjcMSB9Jp\end{tabular}}                                                                                                                                                    & \multicolumn{1}{c|}{0.67}                  & \multicolumn{1}{c|}{24 s}                                                        & \multicolumn{1}{c|}{1}                                                                & \multicolumn{1}{c|}{\begin{tabular}[c]{@{}c@{}}Bitcoin Average - bitcoin \\ price index - (\$ 9638.9) -\\ https://t.co/z6cbnPDdmv\\  \#bitcoin \\ https://t.co/0PoQwUAU1a\end{tabular}}                                                                           & \multicolumn{1}{c|}{0.59}                  & \multicolumn{1}{c|}{0.8s}                                                          & \multicolumn{1}{c|}{1}                                                                \\ \hline
		\multicolumn{1}{|c|}{\begin{tabular}[c]{@{}c@{}}Smart City,\\ 5G, IoT\end{tabular}}                                           & \multicolumn{1}{c|}{\begin{tabular}[c]{@{}c@{}}Samsung IoT Smart City\\ https://t.co/Xnf5JHnOq9\\ via @YouTube \\ @\_funtastic5\_ \\ \#TelkomFuntastic5 \\ \#RWSTREG5 \#smartcity\end{tabular}}                                                                                           & \multicolumn{1}{c|}{0.70}                  & \multicolumn{1}{c|}{25s}                                                         & \multicolumn{1}{c|}{1}                                                                & \multicolumn{1}{c|}{\begin{tabular}[c]{@{}c@{}}Getting Around Smart Cities \\ \#SmartCities via \\ https://t.co/yXaZMpRqm9\\ https://t.co/2HSDlUh7WN\end{tabular}}                                                                                                & \multicolumn{1}{c|}{0.74}                  & \multicolumn{1}{c|}{0.9s}                                                         & \multicolumn{1}{c|}{1}                                                                \\ \hline
		\multicolumn{1}{|c|}{\multirow{3}{*}{\begin{tabular}[c]{@{}c@{}}Joe Biden \\ and \\ Bernie \\ Senders\end{tabular}}}          & \multicolumn{1}{c|}{\begin{tabular}[c]{@{}c@{}}@LyndaMo85130479 \\ @BugOffDear \\ Biden positions are literally just\\ copy/pasted from Bernie Sanders\end{tabular}}                                                                                                                      & \multicolumn{1}{c|}{\multirow{3}{*}{0.46}} & \multicolumn{1}{c|}{\multirow{3}{*}{26 s}}                                       & \multicolumn{1}{c|}{\multirow{3}{*}{3}}                                               & \multicolumn{1}{c|}{\multirow{3}{*}{\begin{tabular}[c]{@{}c@{}}@JoeBiden has become \\ Bernie Sanders 2.0!!!! \\ @JoeBiden\end{tabular}}}                                                                                                                         & \multicolumn{1}{c|}{\multirow{3}{*}{0.81}} & \multicolumn{1}{c|}{\multirow{3}{*}{1.1s}}                                         & \multicolumn{1}{c|}{\multirow{3}{*}{1}}                                               \\ \cline{2-2}
		\multicolumn{1}{|c|}{}                                                                                                        & \multicolumn{1}{c|}{\begin{tabular}[c]{@{}c@{}}Folks mention Biden's past \\ plagiarism True\end{tabular}}                                                                                                                                                                                & \multicolumn{1}{c|}{}                      & \multicolumn{1}{c|}{}                                                            & \multicolumn{1}{c|}{}                                                                 & \multicolumn{1}{c|}{}                                                                                                                                                                                                                                             & \multicolumn{1}{c|}{}                      & \multicolumn{1}{c|}{}                                                            & \multicolumn{1}{c|}{}                                                                 \\ \cline{2-2}
		\multicolumn{1}{|c|}{}                                                                                                        & \multicolumn{1}{c|}{\begin{tabular}[c]{@{}c@{}}But who believes Joe\\ had anything to do \\ with deciding this,\\  or preparing the doc?\\ Who is in charge?\\ https://t.co/x8RBCz4H6D\end{tabular}}                                                                                      & \multicolumn{1}{c|}{}                      & \multicolumn{1}{c|}{}                                                            & \multicolumn{1}{c|}{}                                                                 & \multicolumn{1}{c|}{}                                                                                                                                                                                                                                             & \multicolumn{1}{c|}{}                      & \multicolumn{1}{c|}{}                                                            & \multicolumn{1}{c|}{}                                                                 \\ \hline
		\multicolumn{1}{|c|}{\begin{tabular}[c]{@{}c@{}}Global \\ warming, \\ climate \\ change\end{tabular}}                         & \multicolumn{1}{c|}{So is global warming}                                                                                                                                                                                                                                                 & \multicolumn{1}{c|}{0.67}                  & \multicolumn{1}{c|}{27s}                                                         & \multicolumn{1}{c|}{1}                                                                & \multicolumn{1}{c|}{\begin{tabular}[c]{@{}c@{}}@Ilhan Climate change \\ or global warming?\end{tabular}}                                                                                                                                                          & \multicolumn{1}{c|}{0.95}                  & \multicolumn{1}{c|}{0.9s}                                                          & \multicolumn{1}{c|}{1}                                                                \\ \hline
		\multicolumn{1}{|c|}{\begin{tabular}[c]{@{}c@{}}Food \\ security \\ and \\ Agriculture\end{tabular}}                          & \multicolumn{1}{c|}{\begin{tabular}[c]{@{}c@{}}With food security on the \\ rise, do what you can\\  to help another\\ \#foodsecurity \\ \#food \#endhunger\\  https://t.co/3bLuC6daiN\end{tabular}}                                                                                      & \multicolumn{1}{c|}{0.59}                  & \multicolumn{1}{c|}{26s}                                                         & \multicolumn{1}{c|}{1}                                                                & \multicolumn{1}{c|}{\begin{tabular}[c]{@{}c@{}}RT Moreover Food \\ security and Agricultural \\ self-sufficiency \#foodsecurity\\ \#Agricultural\_sufficiency\\ \#Yemen\end{tabular}}                                                                             & \multicolumn{1}{c|}{0.82}                  & \multicolumn{1}{c|}{0.8s}                                                          & \multicolumn{1}{c|}{1}                                                                \\ \hline
		\multicolumn{1}{|c|}{\begin{tabular}[c]{@{}c@{}}Black lives \\ matter and \\ gunshots\end{tabular}}                                                 & \multicolumn{1}{c|}{\begin{tabular}[c]{@{}c@{}}\#Facebook groups are falling\\  apart over Black Lives\\  Matter posts \\ https://t.co/g0eff0hcLC\\  \#Socialmedia\\  https://t.co/R3dov8nCAq\end{tabular}}                                                                                                   & \multicolumn{1}{c|}{0.59}                                       & \multicolumn{1}{c|}{28s}                                                                              & \multicolumn{1}{c|}{1}                                                                                     & \multicolumn{1}{c|}{\begin{tabular}[c]{@{}c@{}}Black Lives STILL Matter, \\ just in case you forgot. \\ And ALL Lives won’t \\ matter until Black \\ Lives do. \\ https://t.co/bULE7LLNk7\end{tabular} }                                                                               & \multicolumn{1}{c|}{0.68}                                       & \multicolumn{1}{c|}{1s}                                                                               & \multicolumn{1}{c|}{1}                                                                                     \\ \hline
	\end{tabular}
	\caption{Information Foraging Results: EHOIF vs. EEHOLSIF}
	\label{tab6}
\end{table*}

\begin{figure*}
	\centering
	\begin{subfigure}{.4\textwidth}
		\centering
		\includegraphics[scale=0.4]{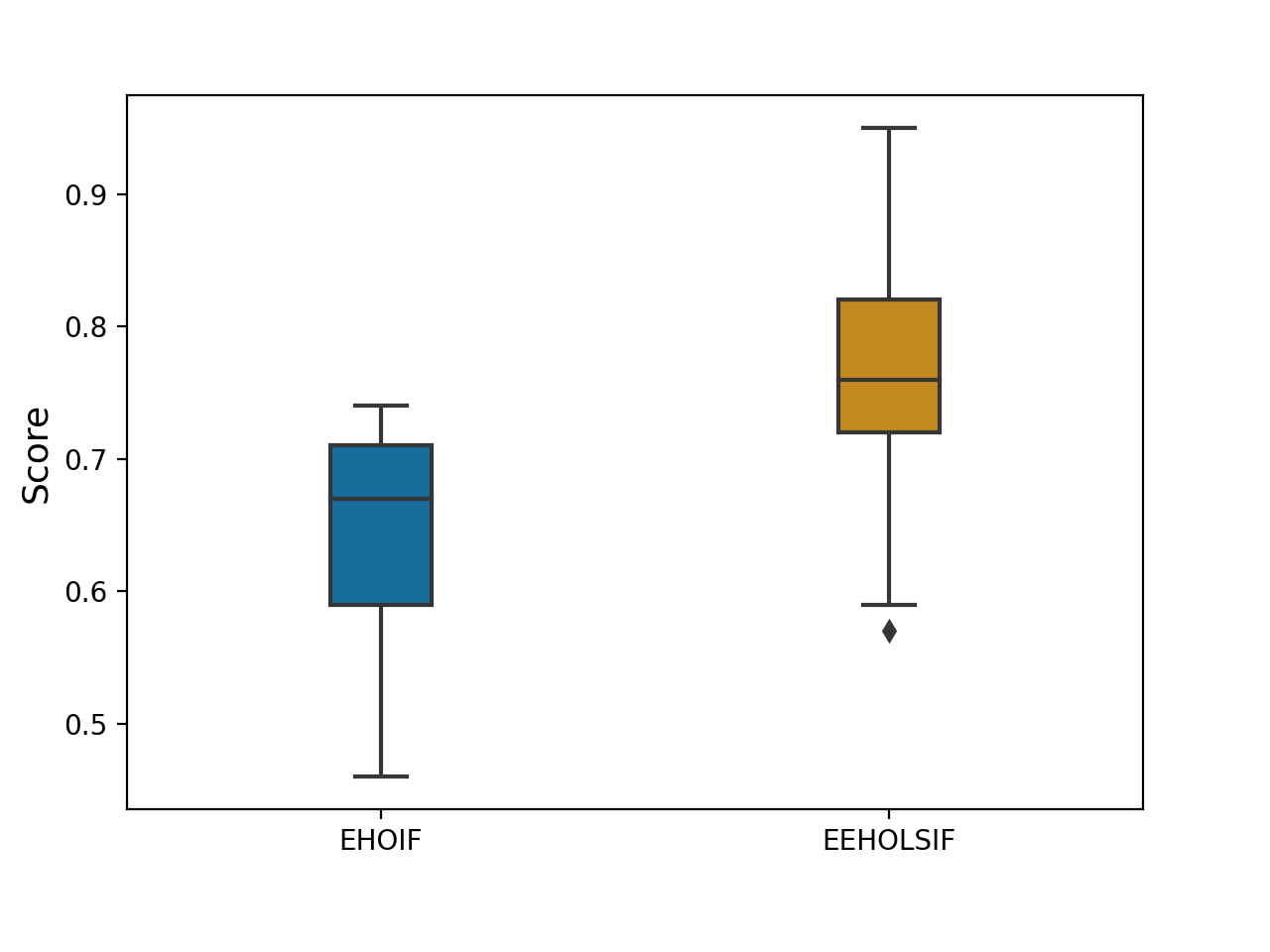}
		\caption{Similarity score}
		\label{fig13a}
	\end{subfigure}%
	\begin{subfigure}{.4\textwidth}
		\centering
		\includegraphics[scale=0.4]{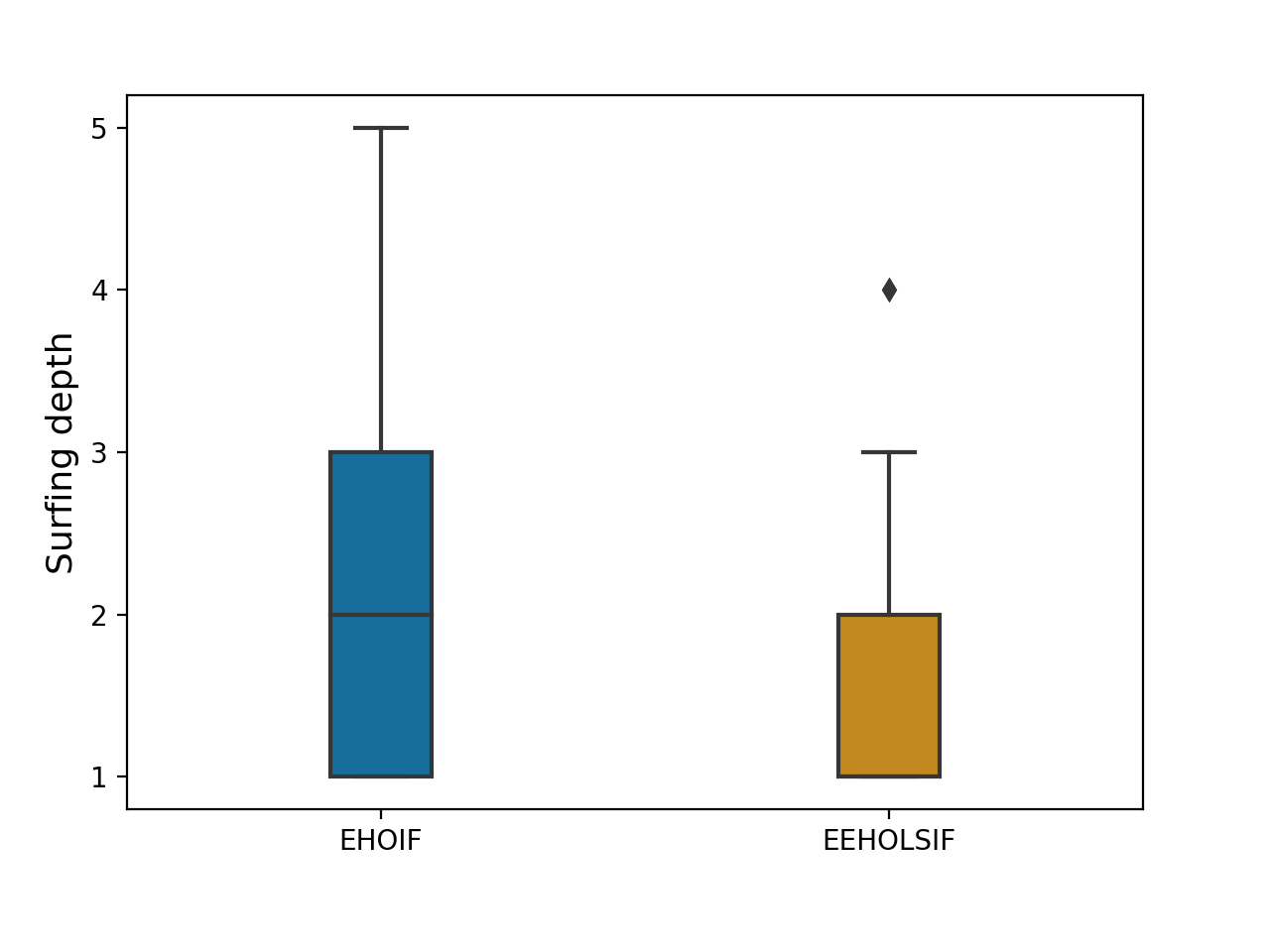}
		\caption{surfing depth}
		\label{fig13b}
	\end{subfigure}
	\begin{subfigure}{.4\textwidth}
		\centering
		\includegraphics[scale=0.4]{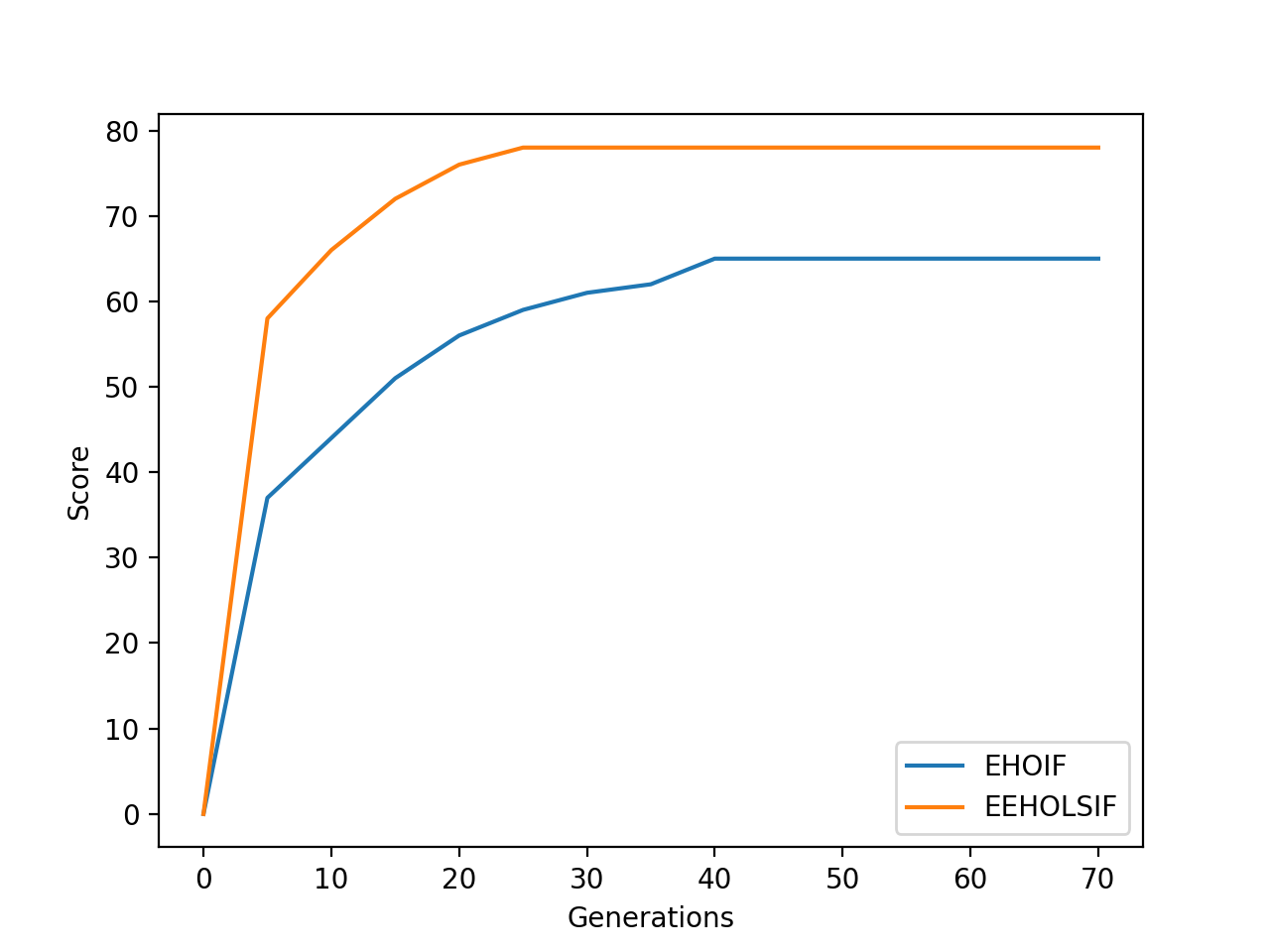}
		\caption{Convergence}
		\label{fig13c}
	\end{subfigure}%
	\begin{subfigure}{.4\textwidth}
		\centering
		\includegraphics[scale=0.4]{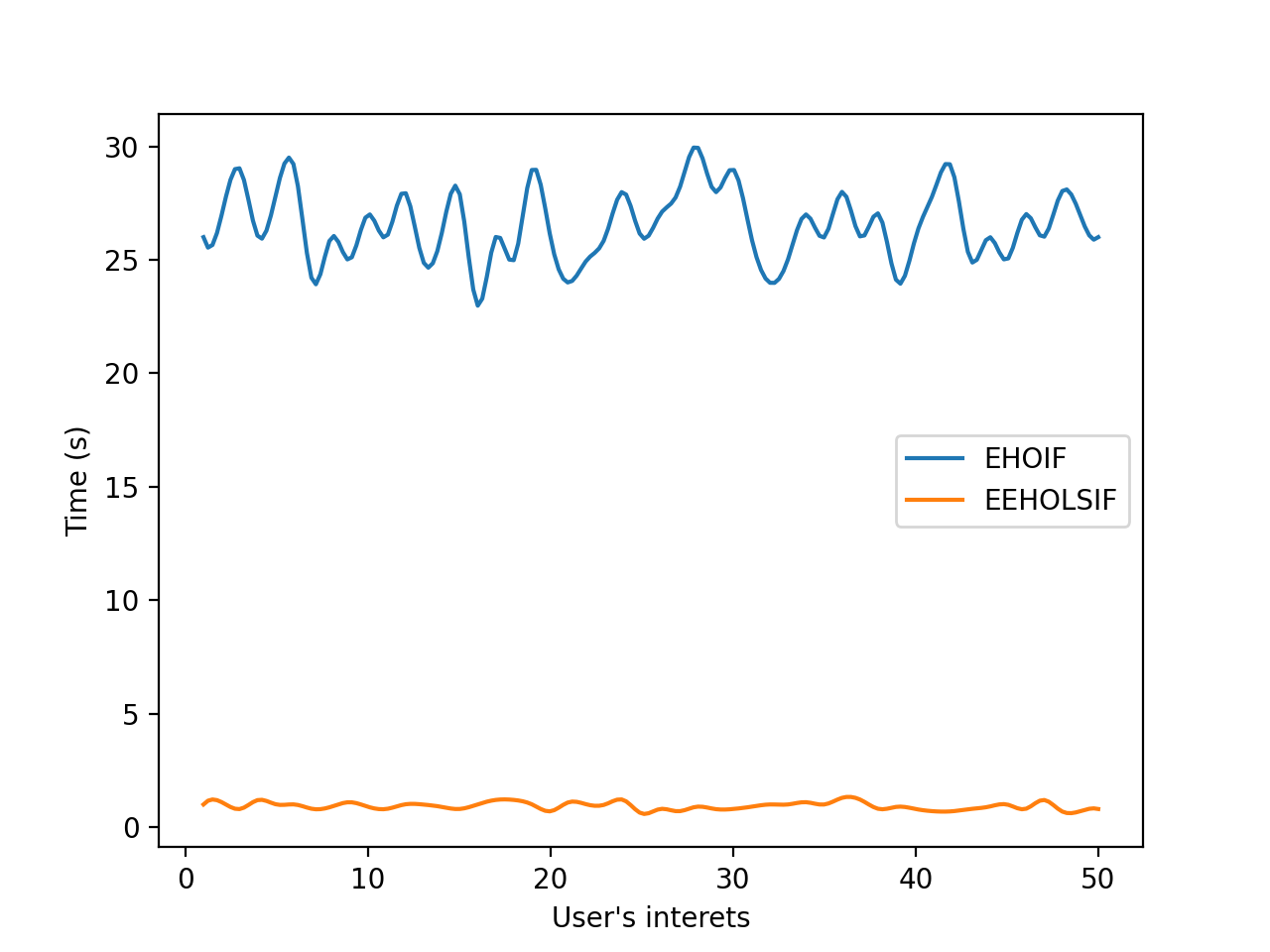}
		\caption{Response time}
		\label{fig13d}
	\end{subfigure}%
	\caption{EHOIF vs. EEHOLSIF}
	\label{fig13}
\end{figure*}

Figure \ref{fig13} displays the comparison results between EHO for information foraging and enhanced EHO for large scale information foraging in terms of relevance score, surfing depth, convergence, and response time. This comparison was made by testing 70 different users' interests, generated randomly, with both approaches. Figure \ref{fig13a} shows that EEHOLSIF is able to achieve better relevance scores with an average of $0.77$ against $0.65$ for EHOIF. Moreover, EEHOLSIF can reach very high scores exceeding $0.9$, while EHOIF is limited to $0.74$. The opposite can be said regarding the surfing depth, which is generally higher with EHOIF as shown in Figure \ref{fig13b}. This is due to the fact that in EHOIF the surfing process is initialized in a complete random way, without taking into consideration the content of the tweets. On the other hand, the territories concept in EEHOLSIF allows to target tweets similar to the user's interest, which helps better guiding the surfing and thus shortening the surfing paths.
As for convergence, we remark that EEHOLSIF has a faster convergence rate while reaching a higher optimum compared to EHOIF. We can see from Figure \ref{fig13c} that EEHOLSIF converges after $25$ generations achieving a relevance score of about $0.77$. EHOIF on the flip side, converges after $40$ generations with a score of $0.65$. This results confirm that the migration mechanism introduced in EEHOLSIF helps preventing stagnation and hence allows the algorithm to reach better solutions rapidly. 
Figure \ref{fig13d} exhibits the run time results, which show that EEHOLSIF is remarkably faster in all cases with an average response time of $0.9$ seconds against $26.5$ seconds for EHOIF. 

\subsection{Comparative study }
A crucial step to validate our work is to compare it to other metaheuristic-based information foraging approaches from the literature. In this section, we pay particular attention to two approaches. 

The first approach we consider is based on \textit{Ant Colony System (ACS)} and was already used to address Web information foraging in \cite{Drias:Kechid:Pasi:16}. First, we implemented ACS and adapted it to tackle information foraging on social media. Then, we conducted a series of tests to set ACS empirical parameters with the aim of maximizing the system's performance. Table \ref{tab7} indicates the parameters' values we fixed following these tests.  

\begin{table}[!htbp]
	\centering
	\begin{tabular}{|c|c|}
		\hline
		\textbf{Parameter} & \textbf{Value} \\
		\hline
		$\alpha$ & 0.2 \\
		\hline
		$\beta$ & 0.4 \\
		\hline
		$\rho$ & 0.8 \\
		\hline
		$q_0$ & 0.8 \\
		\hline
		Number of ants & 50 \\
		\hline
		Number of generations & 50 \\
		\hline
	\end{tabular}
	\caption{ACSIF empirical parameters setting}
	\label{tab7}
\end{table}

The second approach we implemented is based on \textit{Particle Swarm Optimization (PSO)}. To our best knowledge this is the first time PSO is used to address information foraging. Just like with ACS, we carried out several tests to find the optimal values of the empirical parameters, which are presented in Table \ref{tab8}. We denote this approach as PSOIF. 

\begin{table}[!htbp]
	\centering
	\begin{tabular}{|c|c|}
		\hline
		\textbf{Parameter} & \textbf{Value} \\
		\hline
		$c_1$ & 1.5 \\
		\hline
		$c_2$ & 0.4 \\
		\hline
		Number of particles & 600 \\
		\hline
		Number of generations & 90 \\
		\hline
	\end{tabular}
	\caption{PSOIF empirical parameters setting}
	\label{tab8}
\end{table}

\begin{figure*}[!htbp]
	\centering
	\begin{subfigure}{.4\textwidth}
		\centering
		\includegraphics[width=0.9\linewidth]{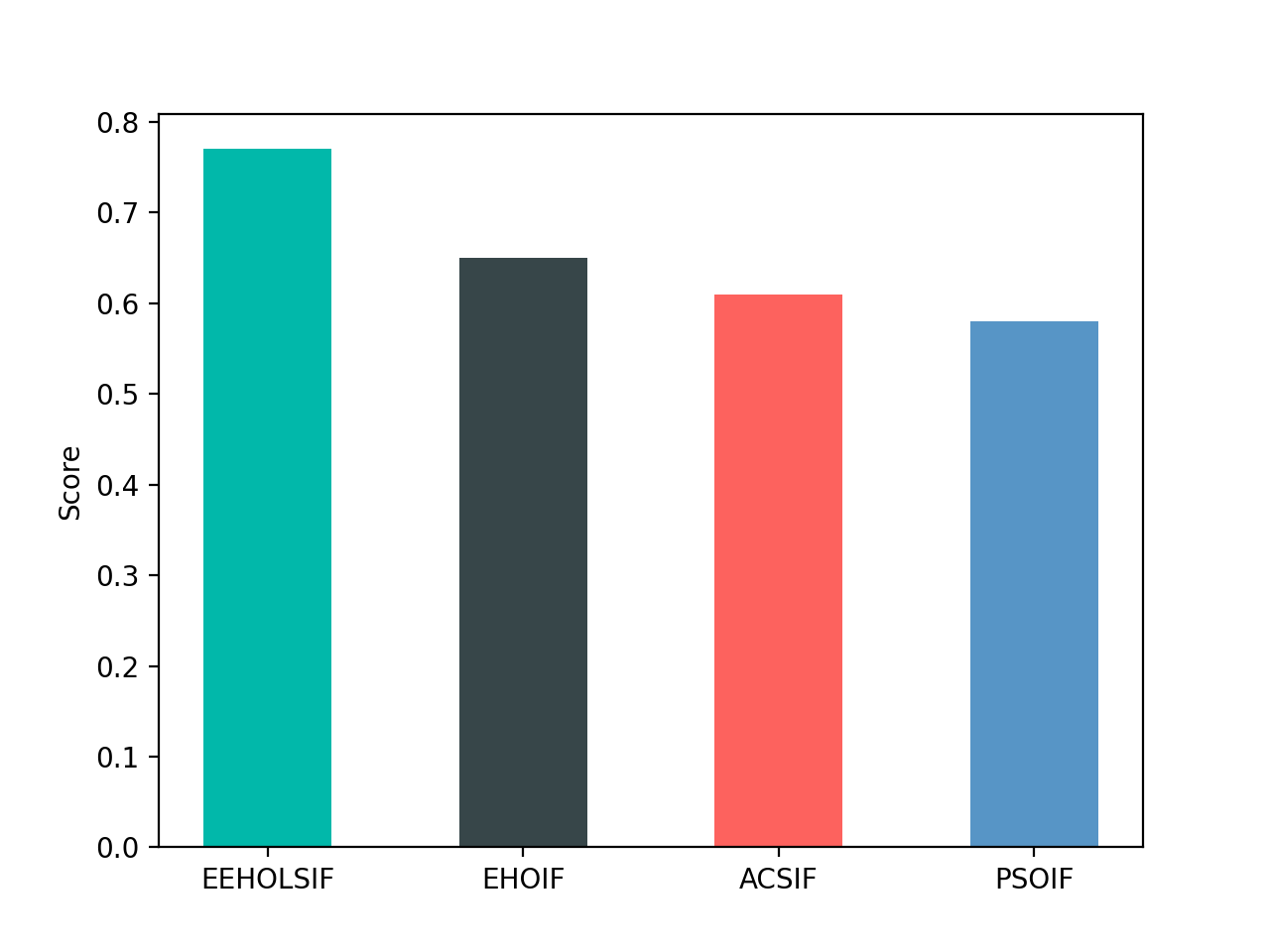}
		\caption{Similarity}
		\label{fig14a}
	\end{subfigure}%
	\begin{subfigure}{.4\textwidth}
		\centering
		\includegraphics[width=0.9\linewidth]{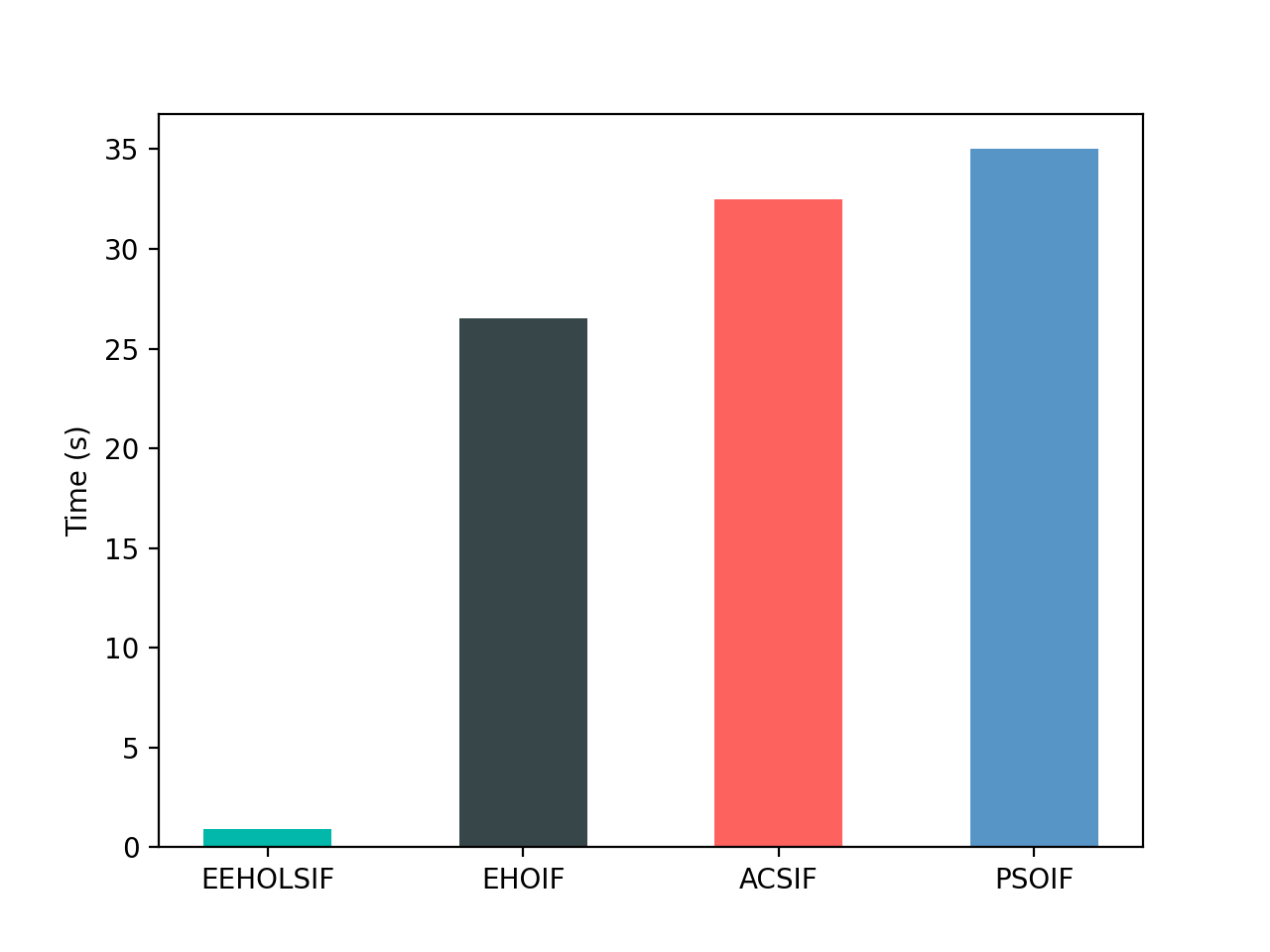}
		\caption{Response time}
		\label{fig14b}
	\end{subfigure}
	\caption{Comparison with ant colony optimization}
	\label{fig14}
\end{figure*}

Once we fixed the empirical parameters for both approaches, we conducted extensive experiments to make a comparative study between EHOIF, EEHOLSIF, ACSIF and PSOIF.  
Figure \ref{fig14a} exhibits the difference in relevance sore between the four approaches. We can see that EEHOLSIF achieves the highest score followed by EHOIF then ACSIF and finally PSOIF. We also notice that the difference in score is quite significant between EEHOLSIF and the other approaches. 
When it comes to response time, Figure \ref{fig14a} shows that EEHOLSIF is the fastest with an average lower than 1 second. The three other approaches have an average response time between 25 and 35 seconds. This means that EEHOLSIF is more than 25 times faster than the other approaches.

We assume that the main problem faced by EHOIF, ACSIF and PSOIF resides in the social graph's size. For instance, ACS is well adapted to work on graphs, since it was first developed to solve the traveling salesman problem. It was also applied to information foraging and gave good results both in terms of score and response time. However, it was only tested on limited size web graphs, which contain less than 2000 web pages \cite{Drias:Kechid:Pasi:16}. Once we increase the number of the web pages or in our case social posts, the complexity of the problem causes a noticeable slow down in terms of response time and a decrease in the relevance score. 

\section{Conclusion and perspectives }\label{sec8}
A novel bio-inspired approach to large scale information foraging using enhanced elephant herding optimization and clustering was proposed in this paper. 
First, we adapted the Elephant Herding Optimization algorithm to information foraging on social media. EHO was originally proposed to solve continuous optimization problems, so to make it able to work on combinatorial problems and more precisely information foraging, we undertook modifications on some important aspects including the elephants' positions implementation, the solution construction and the solution evaluation. To our best knowledge, this is the first attempt to use EHO to address information foraging. The first results show that our approach based on the adaptation of EHO has the ability to find relevant information on social graphs. However, the response time can be very slow especially when the social graph is big. 

To overcome this issue and better handle large scale information foraging, we proposed a new enhanced version of EHO. We introduced several new concepts to the algorithm including two natural phenomena related to elephants' behavior, namely territories delimitation and clan migration. Clustering and more precisely k-means algorithm was used for the implementation of the territories delimitation. Thanks to this phase, we were able to define a better representation of the elephants positions taking into account the real distance between the potential solutions on the search space. Furthermore, dividing the search space into multiple small territories and bounding the foraging to just a few of them, using a newly introduced pseudo random proportional rule, helped to substantially reduce the problem complexity. In addition to that, the clan migration phenomenon prevents the algorithm from stagnation and premature convergence. 

In order to evaluate the proposed approach, we built a dataset containing more that 1.4 million tweets covering different topics. We conducted extensive experiments to test both the adapted EHO for information foraging and the enhanced EHO for large scale information foraging. The results showcase the advantages of EEHOLSIF compared to EHOIF in diffrent aspects including relevance score, response time, convergence and surfing depth. They also demonstrate that the new concepts introduced in EEHOLSIF contribute in boosting the performance considerably.  

Finally, we did a comparative performance analysis with two metaheuristic-based information foraging approaches. The first one being Ant Colony System and the second one being Particle Swarm optimization. The outcomes show that our approach has an much better performance. As far as we know, this is the first time an information foraging approach based on elephant herding optimization is proposed and one of the few destined to work on social media, which gives this work a remarkable originality.

Further work will focus on the parallel implementation of EEHOLSIF using GPUs, as well as a dynamic territories definition process using deep learning for a real time clustering. Another intersting direction would be to integrate a dictionary such as WordNet to better cover the semantic features of the tweets in the vector space representation.     

\section*{Acknowledgement}
This study is funded in part by the General Directorate of Scientific Research and Technological Development (DGRSDT), under Grant No. C0662300.



\vspace{5mm}

\noindent{\bf\Large Author Biographies} \vspace{5mm}

\textbf{Yassine Drias} received his Ph.D. degree in Computer Science from the University of Milano-Bicocca in 2017. Prior to that, he prepared his Master’s degree in Intelligent Informatics Systems at USTHB University, Algeria in collaboration with ENSMA, France and graduated in 2013. He is an Assistant Professor at the University of Algiers. He is currently working on topics including Bio-inspired Computing,Web Information Foraging, Multi-Agent Systems, Machine Learning and Data Mining. His works have appeared in computer science journals and international conferences proceedings.

\vspace{4mm}

\textbf{Habiba Drias} received the M.S. degree in computer science from CWRU Cleveland OHIO USA and the Ph.D. degree in Computer Science from USTHB/Paris6, Algiers. She is a full professor at USTHB University and the head of the Laboratory of Research in Artificial Intelligence (LRIA). She has published more than 200 papers in the domain of artificial intelligence, e-commerce, satisfiability problem, multi-agent systems, meta-heuristics
and large scale information retrieval and data mining in well recognized international conference proceedings and journals. She has also directed 25 Ph.D. theses, 38 master theses and 31 engineer projects. In 2013, she won the Algerian Scopus award in computer science and in 2015, she was selected by a jury of international academicians as a founding member of the Algerian Academy of Science and Technology (AAST).

\vspace{4mm}

\textbf{Ilyes Khennak} received his Ph.D. degree in Computer Science from USTHB University, Algiers in 2017 and the Master's degree in Intelligent Informatics Systems at USTHB University, Algiers in 2011. He is currently an Assistant Professor at USTHB University and his research interests include Metaheuristics, Information Retrieval, Bio-inspired Computing and Data Mining. His works have appeared in computer science journals and international conferences proceedings.

\end{document}